\documentclass{article}

\usepackage{microtype}
\usepackage{graphicx}
\usepackage{subfig}
\usepackage{booktabs} 
\usepackage{amsmath,amssymb,xcolor}
\usepackage{mathtools, bm}
\usepackage{amsthm}

\usepackage{adjustbox}

\theoremstyle{plain}
\newtheorem{theorem}{Theorem}[section]

\newtheorem{lemma}[theorem]{Lemma}

\theoremstyle{definition}
\newtheorem{definition}[theorem]{Definition}

\theoremstyle{remark}

\usepackage{authblk}

\usepackage{hyperref}

\DeclarePairedDelimiterX{\norm2}[1]{||}{||^2_2}{#1}

\usepackage[accepted]{icml2022}
\usepackage[capitalize,noabbrev]{cleveref}

\icmltitlerunning{Efficient Embedding of Semantic Similarity in Control Policies via Entangled Bisimulation}

\begin{document}
\twocolumn[
\icmltitle{Efficient Embedding of Semantic Similarity in Control Policies via Entangled Bisimulation}
\date{\vspace{-5ex}}

\begin{icmlauthorlist}
\icmlauthor{Martin Bertran}{duke,appleintern}
\icmlauthor{Walter Talbott}{apple}
\icmlauthor{Nitish Srivastava}{apple}
\icmlauthor{Joshua Susskind}{apple}
\end{icmlauthorlist}

\icmlaffiliation{duke}{Duke University}
\icmlaffiliation{apple}{Apple}
\icmlaffiliation{appleintern}{Work done as an intern at Apple}

\icmlcorrespondingauthor{Martin Bertran}{martin.bertran@duke.edu}

\icmlkeywords{Machine Learning, ICML}

\vskip 0.3in
]

\printAffiliationsAndNotice{}
\begin{abstract}
Learning generalizeable policies from visual input in the presence of visual distractions is a challenging problem in reinforcement learning. Recently, there has been renewed interest in bisimulation metrics as a tool to address this issue; these metrics can be used to learn representations that are, in principle, invariant to irrelevant distractions by measuring behavioural similarity between states. An accurate, unbiased, and scalable estimation of these metrics has proved elusive in continuous state and action scenarios. We propose \textit{entangled} bisimulation, a bisimulation metric that allows the specification of the distance function between states, and can be estimated without bias in continuous state and action spaces. We show how entangled bisimulation can meaningfully improve over previous methods on the Distracting Control Suite (DCS), even when added on top of data augmentation techniques.
\end{abstract}

\section{Introduction}

Learning control policies from primarily visual input is an important task in many real world applications \cite{lillicrap2015continuous,jaderberg2016reinforcement,kalashnikov2018qt,espeholt2018impala}. While, in principle, reinforcement learning (RL) algorithms can eventually learn a suitable control policy with sufficient experience, poor sample efficiency and the time and monetary cost of data collection still present significant hurdles in practice \cite{duan2016rl, henderson2018deep, kaiser2019model}. Moreover, these agents may additionally suffer from poor generalization to unseen, but semantically equivalent, scenarios \cite{des2018learning,song2019observational,cobbe2019quantifying}, these distractions can also impact learning efficiency.

One promising approach to deal with distractors while also preserving task relevant information is the use of bisimulation metrics \cite{ferns2011bisimulation,ferns2014bisimulation,castro2020scalable}, where states that are indistinguishable with respect to future reward trajectories are grouped together \cite{zhang2020learning} (deep bisimulation for control, DBC), or states which produce the same action sequence under the optimal policy are similarly grouped \cite{agarwal2021contrastive} (policy similarity metrics, PSM). Bisimulation tackles generalization by incorporating and measuring important invariances of the policy and environment dynamics into the representation learning process. Bisimulation metrics have rich theoretical underpinnings \cite{ferns2011bisimulation,ferns2014bisimulation}, but efficient and unbiased estimation of this metric is an active research area; with  current research efforts aimed at stabilizing this metric learning process in practice \cite{kemertas2021towards}.

In this work, we provide a unifying view of currently used bisimulation metrics, their empirical approximations, and provide a mathematically formal way of producing unbiased sample estimates of bisimulation metrics in continuous, stochastic environments for a wide variety of desired invariance specifications. We show how this contribution can meaningfully improve results on current approximations on the Distracting Control Suite benchmark \cite{stone2021distracting}; as well as an experimental analysis on how bisimulation interacts with existing augmentation techniques and agent capacity.

\section{Related Work}

{\color{black} The works in \cite{laskin2020reinforcement,kostrikov2020image} make explicit use of low-level data augmentation to improve the robustness of the learned policy to predefined input perturbations. These works have proven to be very effective at tackling these issues, since the perturbations are pre-specified and can be generated without additional interaction with the environment, improving sample efficiency. One of the drawbacks of these approaches is that the invariances must be known a-priori.

Contrastive learning \cite{srinivas2020curl,chen2020simple,hjelm2018learning} allows the algorithm designer to specify positive and negative matches in representation space, and embed a similarity measure based on this by maximizing agreement between positive matches, and minimizing it w.r.t negative matches; contrastive learning has been shown to improve sample efficiency. Observation prediction and reconstruction \cite{hafner2019dream,sekar2020planning} also provides a rich auxiliary training signal, but forces the agent to model and reconstruct task-irrelevant distractors, which may be a significant disadvantage on natural scenarios \cite{zhang2020learning,agarwal2021contrastive}.}

The work in \cite{ferns2004metrics,ferns2011bisimulation} proposed a bisimulation metric that required the computation of an auxiliary adversarial policy, as well as the computation of a $1-$Wasserstein coupling between state transition distributions, this metric was shown to be a value function of an augmented (bi-state) environment \cite{ferns2014bisimulation}. The work in \cite{castro2020scalable} addressed the overly pessimistic and computationally costly concern of optimizing over an adversarial policy by proposing policy bisimulation, where the policy-averaged rewards and transition functions are used instead. Estimating the $1-$Wasserstein coupling remained a challenging issue, so the work analyzed how to efficiently compute this quantity for deterministic state transitions in Markov decision processes (MDP).

The works of \cite{agarwal2021contrastive, zhang2020learning} (DBC, PSM) tackle the issue of approximating policy bisimulation (or related notions) over partially observable Markov decision processes (POMDP). Both approximations unfortunately suffer from drawbacks where their proposed bisimulation optimization objectives have nontrivial biases for estimating Wasserstein distances, and provably lose important metric properties like state self similarity. Although both have shown to be effective methods in practice, the differences in implementation, methodology and architecture make comparisons between them nontrivial. In this work, we will address these shortcomings, and compare the corrected methods to examine how effective reward- (DBC) or policy-based (PSM) bisimulation metrics are at dealing with semantically equivalent distractions.

\section{Contributions}

\begin{itemize}
    \item We discuss and analyze policy-based bisimulation metrics for continuous action and state spaces and their practical implementations and biases in a unifying framework.
    \item We provide an alternative bisimulation metric, \textit{entangled} bisimulation, which allows for simple, unbiased estimation for \textit{any} distance measure (reward-based, action-based, or otherwise) under mild assumptions. 
    \item We show the effectiveness of the proposed bisimulation measures on the DCS benchmark, and how it improves performance beyond standard image augmentation.
\end{itemize}

\section{Preliminaries}

\subsection{Bisimulation Metrics}

Bisimulation is a principled way of comparing (latent) states and learning or inducing invariances in a policy or environment; it is a distance metric over state pairs. Here we first introduce bisimulation in the context of a finite-time Markov decision process (MDP), described by the tuple $\mathcal{M} \coloneqq (\mathcal{Z}, \mathcal{A}, \mathcal{P}, \mathcal{R}, \gamma)$, where $\mathcal{Z}$ is the latent state space and $\mathcal{A}$ the action space, the distribution $\mathcal{P}(\bm{z}'\mid \bm{z},\bm{a})$  contains the probabilities of transitioning to state $\bm{z}'$ after performing action $\bm{a}$ on state $\bm{z}$, the agent then receives reward $r=\mathcal{R}(\bm{z},\bm{a}) \in \mathbb{R}$. The goal of the agent is to maximize discounted rewards via policy maximization $\max_\pi \mathbb{E}_\mathcal{P}[\sum_{t=1}^\infty \gamma^t R(\bm{z}_t,\bm{a}_t)]$, where actions are sampled according to a learned policy $\bm{a} \sim \pi(\cdot|\bm{z})$. We later show how bisimulation metrics can be applied and efficiently computed in a partially observable Markov decision process (POMDP) where our agent's latent states $\bm{z}$ define a learned MDP.

{\color{black} Bisimulation metrics are specified by three key components:
\begin{itemize}
    \item An equivalence relation $E \in \mathcal{Z}\times \mathcal{Z}$ between states specifying a desired invariance.
    \item A metric transformation $\mathcal{F}: \mathbb{M} \rightarrow \mathbb{M}$ that has a fixed point on a  metric $d:  \mathcal{Z}\times \mathcal{Z} \rightarrow \mathbb{R},\, d \in \mathbb{M}$ satisfying the equivalence relation (i.e., $F(d)=d, \, d(\bm{z}, \bm{z'})=0 \iff (\bm{z}, \bm{z'}) \in E $). We call this metric the bisimulation metric.
    \item For continuous state and action spaces, we additionally require a computable pointwise estimate of the metric transformation, $\widehat{\mathcal{F}}(d): \mathcal{Z}\times \mathcal{Z} \rightarrow \mathbb{R}$ used to learn the bisimulation metric. The methods discussed in this work make use of an auxiliary objective to learn the bisimulation metric of the form \begin{equation}
    J(\bm{z},\bm{z'}) \coloneqq \norm2{d(\bm{z},\bm{z'}) - \widehat{\mathcal{F}}(d)(\bm{z},\bm{z'})}.
    \label{eq:pibisimfixedpointloss}
\end{equation}
\end{itemize}
}
As a motivating example, we consider $\pi$-bisimulation as proposed in \cite{castro2020scalable}, which compares the state and reward dynamics introduced by the policy $\pi$. We use this definition to show the above key concepts in bisimulation. We then observe practical implementations of this bisimulation metric and extract key insights used to design bisimulation metrics  that measure a wide variety of invariances, and how to ensure that the empirical estimates used to learn these metrics in practice are easy to compute, consistent, and unbiased.

Consider the $\pi$-bisimulation relation as proposed in \cite{castro2020scalable}, which compares the state and reward dynamics introduced by the policy $\pi$. Letting $R(\bm{z},\pi)\coloneqq \mathbb{E}_{ \pi(\bm{a}|\bm{z})}R(\bm{z},\bm{a})  $ and $P(\bm{z}_{+}|\bm{z},\pi) \coloneqq \mathbb{E}_{\pi(\bm{a}|\bm{z})} P(\bm{z}_{+}|\bm{z},\bm{a})$ be the policy-averaged reward and state transitions functions respectively, \cite{castro2020scalable} defines a $\pi$-bisimulation relation as follows:

\begin{definition}\cite{castro2020scalable} Given an MDP $\mathcal{M}$, an equivalence relation $E^\pi\subseteq \mathcal{Z}\times \mathcal{Z}$ is a $\bm{\pi}$\textbf{-bisimulation relation} if whenever $(\bm{z}, \bm{z'}) \in E^\pi$ the following properties hold
\begin{equation}
\begin{array}{rl}
     R(\bm{z},\pi) &= R(\bm{z'},\pi)  \\
     P(\cdot|\bm{z},\pi) & =P(\cdot|\bm{z}',\pi).
\end{array}
\label{eq:pibisimrelation}
\end{equation}
\end{definition}

Further, we say that states $\bm{z}, \bm{z'}$ are \textbf{$\bm{\pi}-$bisimilar} if $\bm{z}, \bm{z'}\in  E^\pi$ for some $E^\pi$. We say that $d^\pi \in \mathbb{M}: \mathcal{Z} \times \mathcal{Z} \rightarrow \mathbb{R}^+$ is a \textbf{$\bm{\pi-}$bisimulation metric} if $d(\bm{z},\bm{z'})=0 \leftrightarrow (\bm{z},\bm{z'})$ are $\pi-$bisimilar. Note that this definition implies that two states are identical if both states are expected to collect the same reward throughout every step of the rollout policy $\pi$. 

To find such a bisimulation metric, the work in \cite{castro2020scalable} makes use of a fixed point formulation using the $1-$Wasserstein metric, 

\resizebox{\columnwidth}{!}{\begin{minipage}{\columnwidth}
\begin{equation}
\begin{array}{cc}
     \mathcal{W}_1(P,Q; d) &\coloneqq \inf\limits_{\gamma \in \Gamma(P,Q)}\mathbb{E}_{\bm{z}_+,\bm{z'}_+\sim \gamma}[d(\bm{z}_+,\bm{z'}_+)], \\
\end{array}
\label{eq:Wasserstein}
\end{equation}
\end{minipage}}

\noindent where $P, Q$ are two distributions over the same support, and $\Gamma(P,Q)$ is the set of all couplings of $P,Q$\footnote{joint distribution $\gamma(\bm{z}_+,\bm{z'}_+) $ is a coupling of distributions $P(\bm{z}_+) ,Q(\bm{z}'_+)$, denoted $\gamma \in \Gamma(P,Q)$, if $\int \gamma(\bm{z}_+,\bm{z'}_+) d\bm{z'}_+ = P(\bm{z}_+)$ and $\int \gamma(\bm{z}_+,\bm{z}_+) d\bm{z'}_+ = Q(\bm{z}'_+)$}.

\begin{theorem}\cite{castro2020scalable} Define the metric operator $\mathcal{F}^\pi: \mathbb{M} \rightarrow \mathbb{M}$ as

\resizebox{\linewidth}{!}{
  \begin{minipage}{1.\linewidth} 
\begin{equation}
\begin{array}{rl}
    \mathcal{F}^\pi(d)(\bm{z},\bm{z'}) =& |R(\bm{z},\pi) - R(\bm{z'},\pi)| \\&+ c \mathcal{W}_1(P(\cdot|\bm{z},\pi), P(\cdot|\bm{z}',\pi); d).
\end{array}
\label{eq:pibisimtheo}
\end{equation}
\end{minipage}}

Then, with $c \in [0,1)$, $\mathcal{F}^\pi$ has at least a fixed point $d_\pi$ which is a bisimulation metric.
\end{theorem}

The Wassertein metric is used to optimally compare differences in state transitions, and plays a vital role in ensuring that states are self-similar ($d_\pi(\bm{z},\bm{z})=0\, \forall\, \bm{z}$). That is, bisimulation measures differences in results attributed to the agent's actions in the environment and its current latent state, and optimally discounts the environment's subsequent stochastic transitions. Without this Wasserstein coupling, distances between states would be inflated due to environment stochasticity\footnote{{\color{black}for example, if both $P$ and $P'$ were equally likely to transition to the same pair of dissimilar states $\bm{z}_1, \bm{z_2}$ we would have $\mathbb{E}_{\bm{z}_+, \bm{z}'_+\sim P\times P'}d(\bm{z}_+,\bm{z'}_+)=\frac{1}{2}d(\bm{z}_1, \bm{z_2})$, while $\mathcal{W}_1(P,P'; d)=0$}}, and, most crucially, the ordering of distances between states would no longer encode potential behavioural differences.

We now look at the propsed computable approximations of $\mathcal{F}^\pi(d)(\bm{z},\bm{z'})$. This poses a significant challenge since it requires the computation of the policy-averaged state transition $P(\bm{z}_{+}|\bm{z},\pi)$, and, most importantly, $\mathcal{W}_1(P,Q; d)$ over a multivariate continuous distribution, which is a notoriously hard estimation problem in general \cite{chizat2020faster}.

Two recent approaches to embed $\pi-$bisimulation metrics are DBC \cite{zhang2020learning} and PSM \cite{agarwal2021contrastive}. We first analyze these approaches before presenting our solution.  

The work in \cite{zhang2020learning} (DBC) uses the following relaxation for bisimulation estimation, which uses the $\mathcal{W}_{2}$ distance
\resizebox{\linewidth}{!}{
  \begin{minipage}{1.\linewidth} 
\begin{equation}
    \begin{array}{rl}
d_{DBC}(\bm{z},\bm{z}') &\coloneqq ||\bm{z}-\bm{z}'||_1\\
\mathcal{F}_{DBC}(d)(\bm{z},\bm{z'})&\coloneqq |R(\bm{z},\pi)-R(\bm{z}',\pi)|  \\
&+ c\mathcal{W}_{\color{blue}2}(P(\bm{z}_{+}|\bm{z},\pi), P(\bm{z}'_{+}|\bm{z}',\pi); \ell_1)],\\
\widehat{\mathcal{F}}_{DBC}(d)(\bm{z},\bm{z'})&\coloneqq |R(\bm{z},\bm{a})-R(\bm{z}',\bm{a}')|  \\
&+ c \mathcal{W}_{\color{blue}2}(P(\bm{z}_{+}|\bm{z},\bm{a}), P(\bm{z}'_{+}|\bm{z}',\bm{a}'); \ell_1),\\
\bm{a},\bm{a'} &\sim  \pi(\cdot|\bm{z})\times\pi(\cdot|\bm{z}').
    \end{array}
    \label{eq:dbcobj}
\end{equation}
\end{minipage}}

Where they define the bisimulation metric to be the $\ell_1$ distance between states, learn the latent state dynamics $P$ to conform to $d_{DBC}(\bm{z},\bm{z}') = \mathcal{F}_{DBC}(d)(\bm{z},\bm{z'})$, and plug in single-sample estimates of $R(\bm{z},\pi), R(\bm{z}',\pi), P(\bm{z}_{+}|\bm{z},\bm{a}),$ and $P(\bm{z}_{+}|\bm{z}',\bm{a})$, with actions $(\bm{a},\bm{a'})$ independently sampled from each policy $\bm{a},\bm{a'} \sim  \pi(\cdot|\bm{z})\times\pi(\cdot|\bm{z}')$ to create their computable approximation $\widehat{\mathcal{F}}_{DBC}(d)(\bm{z},\bm{z'})$. The key insight is that $\mathcal{W}_{2}(P,Q;\ell_1)$ can be computed in closed form if $P,Q$ are multivariate Gaussian distributions, so the authors make use of this.

The specific choice of the distance function links the overall scale of latent states to the environment reward, since doubling the reward function doubles the average distance between states in the DBC formulation. A more pressing issue is that the proposed relaxation does not satisfy self similarity. That is,  $\widehat{\mathcal{F}}_{DBC}(d_{DBC})(\bm{z},\bm{z})>0$, since
\begin{equation}
    \mathbb{E}_{\pi(\bm{a}|\bm{z})\times\pi(\bm{a}'|\bm{z})}||R(\bm{z},\bm{a})-R(\bm{z},\bm{a}')||_1>0,
\end{equation}
unless $R(\bm{z},\bm{a}) =R(\bm{z},\bm{a'})\, \forall \bm{a},\bm{a'}$ in the support of $\pi(\cdot|\bm{z})$, and, likewise

\resizebox{\linewidth}{!}{
  \begin{minipage}{1.\linewidth} 
\begin{equation}
    \mathbb{E}_{\bm{a},\bm{a'}\sim \pi(\cdot|\bm{z})\times\pi(\cdot|\bm{z})}[\mathcal{W}_{2}(P(\cdot|\bm{z},\bm{a}), P(\cdot|\bm{z},\bm{a}'); ||\cdot||_1)]>0.
\end{equation}
\end{minipage}}

The choice of the pointwise approximation $\widehat{\mathcal{F}}_{DBC}(d)$ means that there is no metric $d$ and transition function $P$ minimizing Eq. \ref{eq:pibisimfixedpointloss} that satisfies the fixed point relation.

PSM proposes a related bisimilarity metric where they compare the average action of the optimal policy in place of average reward comparison\footnote{We specifically discuss their formulation for continuous action and state spaces}
\resizebox{\linewidth}{!}{
  \begin{minipage}{1.\linewidth} 
\begin{equation}
 \begin{array}{rl}
 \mathcal{F}_{PSM}(d)(\bm{z},\bm{z'})\coloneqq& ||\mathbb{E}_{ \pi(\bm{a}|\bm{z})}[\bm{a}] - \mathbb{E}_{\pi(\bm{a}'|\bm{z}')}[\bm{a}']||_1\\  &+c \mathcal{W}_1(P(\cdot|\bm{z},\pi), P(\cdot|\bm{z}',\pi); d),\\
 \widehat{\mathcal{F}}_{PSM}(d)(\bm{z},\bm{z'})\coloneqq& ||\mathbb{E}_{ \pi(\bm{a}|\bm{z})}[\bm{a}] - \mathbb{E}_{\pi(\bm{a}'|\bm{z}')}[\bm{a}']||_1\\&+c d(\bm{z}_+,\bm{z}'_+),\\
 \bm{a},\bm{a'} \sim &  \pi(\cdot|\bm{z})\times\pi(\cdot|\bm{z}'),\\
 \bm{z}_+,\bm{z'}_+ \sim &  P(\cdot|\bm{z}, \bm{a})\times P(\cdot|\bm{z}', \bm{a}').
    \end{array}
\label{eq:psebisim}
\end{equation}
\end{minipage}}

We note that the expected policy action $\mathbb{E}_{\pi(\bm{a}|\bm{z})}[\bm{a}]$ is explicitly computed in most agent architectures, so there is no estimation error with this part of the estimate. The Wasserstein distance is replaced by an expectation over independent state transition and policy distributions; this allows for easy computation, but does not consider all potential couplings as in the Wasserstein distance. Similarly to DBC, the proposed relaxation does not satisfy self-similarity, that is, $\widehat{\mathcal{F}}_{PSM}(d_{PSM})(\bm{z},\bm{z})>0$ since $\mathbb{E}_{\substack{\bm{a},\bm{a'} \sim  \pi(\cdot|\bm{z})\times\pi(\cdot|\bm{z}'),\\\bm{z}_+,\bm{z'}_+ \sim  P(\cdot|\bm{z}, \bm{a})\times P(\cdot|\bm{z}', \bm{a}')}}d_{PSM}(\bm{z}_+,\bm{z}'_+) >0$ unless all independently reachable states $\bm{z}_+,\bm{z'}_+$ have $0$ distance. Since self-similarity is not satisfied, then we can conclude that $d_{PSM}$ is not a metric.

\subsection{Entangled Bisimulation}

We introduce $\epsilon$-bisimulation (entangled bisimulation) which allows for simple, principled computation in standard RL settings without approximation bias. We show how this formulation relates to the DBC and PSM measures discussed in the previous paragraphs with minimal adaptations, while still showing the desireable properties of a proper bisimulation metric. To achieve unbiased estimation, we first introduce entangled sampling.

\begin{definition} Given two multivariate distributions $P,Q$ defined over $\mathbb{R}^n$, and let $F_P^{-1}, F_Q^{-1} $ be the inverse sampling functions of $P$ and $Q$ respectively\footnote{e.g., for a two-dimensional vector $\{x_1,x_2\}\sim P$ we have $F_P^{-1}(u_1,u_2)= F_{P_{x_1}}^{-1}(u_1),F_{P_{x_2|x_1}}^{-1}(u_2)$ } the \textbf{entangled coupling} of $P,Q$, denoted $\gamma^\epsilon(P,Q)$ is defined as:
\begin{equation}
    \begin{array}{rl}
         x, y =& F_P^{-1}(\bm{\epsilon}),F_Q^{-1}(\bm{\epsilon}), \; \bm{\epsilon} \sim U_{[0,1]}^{\otimes n},\\
         x,y & \stackrel{d}{\sim} \gamma^\epsilon(P,Q).
    \end{array}
    \label{eq:entangledcoupling}
\end{equation}
\end{definition}

Note that for architectures using the reparametrization trick,\footnote{for example, to take a continuous action sample in the $[-1,1]$ range with hyperbolic tangent squashing  $\bm{a} \sim \pi(\bm{a}\mid \bm{z})$, we compute state dependent mean and standard deviations  $\mu_i(\bm{z}), \sigma_i(\bm{z})$, and sample $a_i = \tanh(\mu_i(\bm{z}) + \epsilon_i \sigma_i(\bm{z})), \epsilon_i \sim \mathcal{N}(0,1)$}, the entangled coupling amounts to reusing the noise vector between two distributions. To improve readability, we use a shortened notation to denote two couplings of interest, namely policy and state couplings, as:\begin{equation}
        \begin{array}{rl}
        \gamma^\epsilon_\pi =&\gamma^\epsilon(\pi(\cdot|\bm{z}),\pi(\cdot|\bm{z}')),\\
        \gamma^\epsilon_P =&\gamma^\epsilon(P(\cdot|\bm{z},\bm{a}),P(\cdot|\bm{z}',\bm{a}')).
        \end{array}
\end{equation} 
Where we omit the state conditioning on the policy coupling, and the state action conditioning on the state coupling. We define a family of equivalence relations based on any state-action metric $G: \!(\mathcal{Z}\!\times\!\mathcal{A})^2\rightarrow \!\mathbb{R}^+\!, G\in \!\mathbb{M}$:

 \begin{definition}Given an MDP $\mathcal{M}$, an equivalence relation $E^\epsilon\subseteq \mathcal{Z}\times \mathcal{Z}$ is a $\bm{\epsilon}$\textbf{-bisimulation relation} (entangled bisimulation relation) w.r.t policy $\pi$ and state-action metric $G$ if whenever $(\bm{z}, \bm{z'}) \in E^\epsilon$ the following properties hold $\forall \bm{a},\bm{a}' \in \textit{Supp}(\gamma^\epsilon_\pi)$
 \label{def:entangledbisimrelation}
\begin{equation}
\begin{array}{rl}
     G(\bm{z},\bm{a},\bm{z}',\bm{a}') &=0  \\
     P(\cdot|\bm{z},\bm{a})  &=P(\cdot|\bm{z}',\bm{a}').
\end{array}
\label{eq:entangledbisimrelation}
\end{equation}
\end{definition}

Where we require $G$ to satisfy all core metric properties, that is
\begin{definition} A function $G: (\mathcal{Z}\times\mathcal{A})^2\rightarrow \mathbb{R}^+$ is a \textbf{state-action similarity metric} if $\forall \bm{z},\bm{z}',\bm{z}'',\bm{a},\bm{a}',\bm{a}'' \in \mathcal{Z}^3\times\mathcal{A}^3$, $G$ satisfies
\resizebox{\linewidth}{!}{
  \begin{minipage}{1.\linewidth} 
\begin{equation}
    \begin{array}{lll}
      \text{Non-negativity}:& \!\!\!\! G(\bm{z},\bm{a},\bm{z}',\bm{a}')&\ge 0 \\
      \text{Self-similarity}:& \!\!\!\!  G(\bm{z},\bm{a},\bm{z},\bm{a})&=0 \\
      \text{Symmetry}:&  \!\!\!\! G(\bm{z},\bm{a},\bm{z}',\bm{a}')& = G(\bm{z}',\bm{a}',\bm{z},\bm{a})  \\
      \text{Triangle}&  \!\!\!\! G(\bm{z},\bm{a},\bm{z}'',\bm{a}'')&\le  G(\bm{z},\bm{a},\bm{z}',\bm{a}')\\ \text{Inequality}:&&+ G(\bm{z}',\bm{a}',\bm{z}'',\bm{a}'').\\
    \end{array}
    \label{eq:similarityFunction}
\end{equation}
\end{minipage}}
\end{definition}

We can thus use any state-action similarity metric as a basis for an equivalence class captured by $\epsilon$-bisimulation.We can now compute the associated $\bm{\epsilon}$\textbf{-bisimulation metric} by using the following fixed point theorem
\begin{theorem} Define the metric operator $\mathcal{F}^\epsilon: \mathbb{M} \rightarrow \mathbb{M}$ as 
\label{theofixedpoint}
\resizebox{\linewidth}{!}{
  \begin{minipage}{1.12\linewidth}
\begin{equation}
\begin{array}{rl}
\mathcal{F}_\epsilon(d)(\bm{z},\bm{z'}) =& \mathop{\mathbb{E}}\limits_{\bm{a},\bm{a}'\sim \gamma^\epsilon_\pi} [G(\bm{z},\bm{a},\bm{z}',\bm{a}') \\&+ c \mathcal{W}_1(P(\bm{z}_{+}|\bm{z},\bm{a}), P(\bm{z}'_{+}|\bm{z}',\bm{a}'); d)].
\end{array}
\label{eq:epsilonbisimtheo}
\end{equation}
\end{minipage}}

Then, with $c \in [0,1)$, $\mathcal{F}_\epsilon$ has at least a fixed point $d_\epsilon$ which is an$\epsilon$-bisimulation metric.
\end{theorem}

We observe that entangled bisimulation can build stricter equivalence relations than policy-based bisimulation.

\begin{lemma} Let $E^\pi$ be the largest $\pi$-bisimulation relation, and let $E^{PSM}$ likewise be average action bisimulation relation in \cite{agarwal2021contrastive}. The entangled bisimulation relation $E^\epsilon$ satisfies
\label{lemma:otherdefinitions}
\begin{equation}
\begin{array}{rl}
     E^\epsilon &\subseteq E^\pi  \\
     E^\epsilon &\subseteq E^{PSM}  \\
\end{array}
\end{equation}
for the choice respective choice of state-action metrics 
\begin{equation}
    \begin{array}{rl}
         G(\bm{z},\bm{a},\bm{z}',\bm{a}') &= |R(\bm{z},\bm{a})-R(\bm{z}',\bm{a}')|,\\
         G(\bm{z},\bm{a},\bm{z}',\bm{a}') &=  ||\mathbb{E}_{\pi(\bm{a}|\bm{z})}[\bm{a}]-\mathbb{E}_{\pi(\bm{a}'|\bm{z}')}[\bm{a}']||_1 .
    \end{array}
\end{equation}
\end{lemma}

Finally, to address estimation of the $\mathcal{W}_1$ metric, the following theorem provides an entangled upper bound that still provably satisfies the $\epsilon$-bisimulation relation; this upper bound can be computed as a simple average. Further, when state transition dynamics also satisfy coordinate independence (a common feature in RL architectures), and we restrict the learned distance functions to be of the form $d(\bm{z},\bm{z}')=\sum_i w_i d_i(z_i,z'_i)$, where $w_i \ge 0 $, and $d_i(\cdot,\cdot)$ convex, then the entangled upper bound is exact.

\begin{theorem} Given an MDP $\mathcal{M}$, policy $\pi$, and state-action metric $G$. Define the metric operator $\mathcal{F}_{\bar{\epsilon}}\!: \!\mathbb{M} \!\rightarrow \!\mathbb{M}$ as 
\label{theoMain}
\resizebox{\linewidth}{!}{
  \begin{minipage}{1.\linewidth} 
\begin{equation}
\begin{array}{rl}
    \mathcal{F}_{\bar{\epsilon}}(d)(\bm{z},\bm{z'}) =&\hspace{-8pt}\mathop{\mathbb{E}}\limits_{\substack{\bm{a}, \bm{a'}\sim\gamma^\epsilon_\pi\\\bm{z}_+,\bm{z}'_+\sim \gamma^\epsilon_P}}[G(\bm{z},\bm{a},\bm{z}',\bm{a}') +c d(\bm{z}_{+}, \bm{z}'_{+})].
\end{array}
\label{eq:epsilonuboperator}
\end{equation}
\end{minipage}}

Then, $\mathcal{F}_{\bar{\epsilon}}$ has at least a fixed point $d_{\bar{\epsilon}}$ satisfying $d_{\bar{\epsilon}}\ge d_{\epsilon}$, and $ (\bm{z}, \bm{z'}) \in E^\epsilon \rightarrow d_{\bar{\epsilon}}(\bm{z}, \bm{z'})=0$, in particular, $d_{\bar{\epsilon}}(\bm{z}, \bm{z})=0, \, \forall \bm{z}\in \mathcal{Z}$.

Further, if the state transition function is coordinate independent ($P(\bm{z}_+\mid \bm{z},\bm{a}) =\Pi_i P_i(z_{i,+}\mid\bm{z},\bm{a})\, \forall \bm{z}_+,\bm{z},\bm{a} $), and $d_{\bar{\epsilon}}$ is of the form 

\resizebox{\linewidth}{!}{
  \begin{minipage}{1\linewidth} 
\begin{equation}
    d_{\bar{\epsilon}}(\bm{z},\bm{z}') = \sum\limits_{\substack{i=[n]\\j=[p]}}w_{i,j}|z_i-z'_i|^j, \; w_{i,j} \ge 0 \forall i,j, \; p> 0,
\end{equation}
\end{minipage}}

then the bound is tight, that is

\resizebox{\linewidth}{!}{
\begin{minipage}{1.\linewidth} 
\begin{equation}
    \begin{array}{rl}
     d_\epsilon(\bm{z},\bm{z}') & =\hspace{-5pt}\mathop{\mathbb{E}}\limits_{\substack{\bm{a}, \bm{a'}\sim\gamma^\epsilon_\pi\\\bm{z}_+,\bm{z}'_+\sim \gamma^\epsilon_P
     }}[G(\bm{z},\bm{a},\bm{z}',\bm{a}')+c d_\epsilon(\bm{z}_{+}, \bm{z}'_{+})].
    \end{array}
    \label{eqExpectedBisimTight}
\end{equation}
\end{minipage}}
\end{theorem}

Theorem \ref{theoMain} suggests the use of the sample estimate in Eq.\ref{eqExpectedBisimTight} to estimate bisimilarity embeddings in the same way one would estimate a simple value learning objective,

\begin{equation}
\begin{array}{rl}
 \bm{a},\bm{a'} \sim \gamma^\epsilon_\pi, &
 \bm{z}_+,\bm{z'}_+ \sim  \gamma^\epsilon_P, \\   
     \widehat{\mathcal{F}}_\epsilon(d)(\bm{z},\bm{z'})& = G(\bm{z},\bm{a},\bm{z}',\bm{a}') +c d(\bm{z}_{+}, \bm{z}'_{+})  \\
    J_\epsilon(\bm{z},\bm{z}')) &= ||d(\bm{z},\bm{z}') -\widehat{\mathcal{F}}_\epsilon(d)(\bm{z},\bm{z'})||^2_2.
\end{array}
\label{eqBisimulationLoss}
\end{equation}

Unlike other approaches, which use actions and/or state transitions from the replay buffer, we sample actions and states from our entangled distributions. We note that sampling multiple latent states from the entangled policy is computationally efficient since it requires a single feedforward pass through the network; multi-sample estimates are likewise supported. Note that the entangled policy $\gamma^\epsilon(\pi(\cdot|\bm{z}),\pi(\cdot|\bm{z}'))$ satisfies $\gamma^\epsilon(\pi(\cdot|\bm{z}),\pi(\cdot|\bm{z})))(\bm{a}, \bm{a'}) = \pi(\bm{a}|\bm{z}) \delta_{\bm{a}=\bm{a}'}$ and is smoothly varying with $\pi$. Entangled sampling thus ensures that
\begin{equation}
\mathop{\mathbb{E}}\limits_{\bm{a},\bm{a}'\sim \gamma^\epsilon_\pi} G(\bm{z},\bm{a},\bm{z},\bm{a}') =0\, \forall \pi, \bm{z}, G,
\label{eq:selfsim}
\end{equation}
this property is not shared by independent action sampling\footnote{for continuous variables in general, entangled coupling is a one to one mapping between distributions; for finite, discrete distributions with cardinality $|\mathcal{A}|$, entangled coupling visits at most $2|\mathcal{A}|$ distinct value pairs}. Equation \ref{eq:selfsim} shows that self-similarity is an intrinsic property of this estimator, indeed, we observe that every sample of the distribution satisfies $\widehat{\mathcal{F}}_\epsilon(d)(\bm{z},\bm{z})=0$.

\section{Model description}

Our bisimulation state regularization technique can be added as an extra objective on top of any RL agent that has a latent transition model and a continuous action policy that makes use of reparametrization. To benchmark this regularization, we use SAC \cite{haarnoja2018soft} as the basis for our experimental setup, with minor modifications described next. For some of the experiments, we also consider the use of the data augmentation technique proposed in DrQ \cite{kostrikov2020image}, a state of the art augmentation technique in RL. The model components are:

\begin{equation*}
    \begin{array}{rl}
        \text{Observation encoder} &\bm{z}_t = h_\theta(o_t)\\
        \text{Reward predictor} & \hat{r}_t = R_\theta(\bm{z}_t)  \\
        \text{Inverse dynamics model} & \hat{\bm{a}}_t = A_\theta(\bm{z}_t,\bm{z}_{t-1})  \\
        \text{Transition model} & P_\theta(\hat{\bm{z}}_t|\bm{z}_{t-1},\bm{a}_{t-1})  \\
        \text{Critics} & Q^i_{\theta}(\bm{z}_t,\bm{a}_t),\; i\in \{1,2\}  \\
        \text{Policy} & \pi_\theta(\bm{a}_t|\bm{z}_t)\\
        \text{Bisimulation distance} & d_\theta(\bm{z},\bm{z}')
    \end{array}
\end{equation*}

Algorithm \ref{alg:coreBisim} shows the core $\epsilon$-bisimulation computation, where we use $\overline{\bm{z}}$ to denote the gradient stop operation applied to variable $\bm{z}$, the full optimization loss for the model parameters is computed as \begin{equation}
  J_{SAC} + J_{ID} + \beta J_\epsilon.  
\end{equation}
Here $J_{SAC}, J_{ID}$ and $J_\epsilon$ represent the standard SAC loss, inverse dynamics loss, and the proposed $\epsilon-$bisimulation loss, with parameter $\beta$ controlling the relative importance of the latter objective. The full algorithm is shown in Appendix \ref{sec:algorithm_supplementary}.

\begin{algorithm}
\caption{Bisimulation Algorithm}\label{alg:coreBisim}
\begin{algorithmic}
\REQUIRE Latent state batch $B=\{\bm{z}\}_{i=1}^n$, policy $\pi_\theta$, latent transition model $P_\theta(\cdot\mid,\bm{z},\bm{a})$, bisimulation distance function $d_\theta(\cdot,\cdot)$, similarity pseudometric $G$
\STATE Permute states
\STATE $B'=\{\bm{z}'\} = \text{Perm}(\{\bm{z}\})$
\STATE Sample noise variables 
\STATE$\epsilon^A \sim \mathcal{N}(0, I^{|A|}), \epsilon^Z \sim \mathcal{N}(0, I^{|Z|})$
\STATE Compute tied actions 
\STATE$\bm{a} = \pi_\theta(\cdot|\bm{z}, \epsilon^A),\;  \bm{a}'=\pi_\theta(\cdot|\bm{z}', \epsilon^A)$
\STATE Compute tied latent transitions 
\STATE$\bm{z}_+ = P_\theta(\cdot|\bm{z}, \bm{a},\epsilon^Z),\;  \bm{z}'_+=P_\theta(\cdot|\bm{z}', \bm{a}',\epsilon^Z)$
\STATE Compute bisimulation target
\STATE$\hat{d}(\bm{z},\bm{z}') = G(\bm{z},\bm{a},\bm{z}',\bm{a}') +c d_\theta(\bm{z}_+,\bm{z}'_+)$
\STATE Compute bisimulation loss
\STATE $J_\epsilon(\bm{z},\bm{z}') =\frac{1}{n}\sum\limits_{\bm{z'},\bm{z} \in, B,B'} ||d_\theta(\bm{z},\bm{z}') - \overline{\hat{d}(\bm{z},\bm{z}')}||^2_2$ 
\STATE \OUTPUT $J_\epsilon(\bm{z},\bm{z}')$
\end{algorithmic}
\end{algorithm}

\section{Experiments and Results}

\begin{figure}[h!]
\begin{center}
\footnotesize
\subfloat[][Ball in Cup Catch]{
\includegraphics[trim=0 60 170 60,clip,width=\columnwidth]{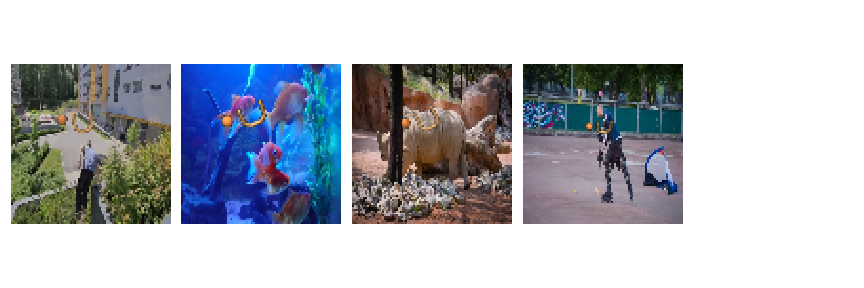}
\label{fig:subfigBIC}}

\subfloat[][Cartpole Swingup]{
\includegraphics[trim=0 60 170 60,clip,width=\columnwidth]{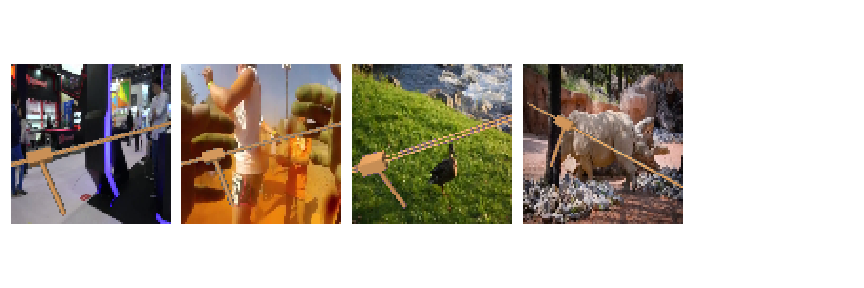}
\label{fig:subfigCS}}

\subfloat[][Cheetah Run]{
\includegraphics[trim=0 60 170 60,clip,width=\columnwidth]{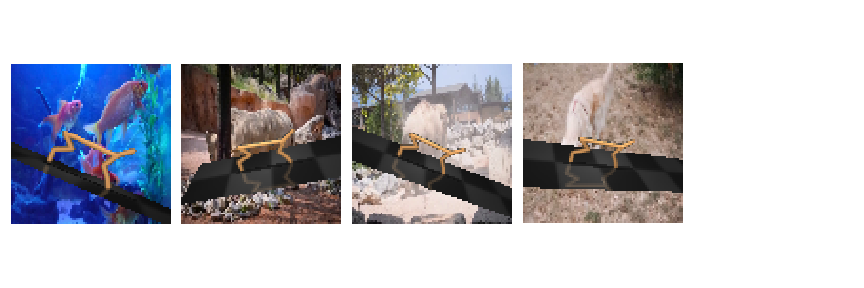}
\label{fig:subfigCR}}

\subfloat[][Finger Spin]{
\includegraphics[trim=0 60 170 60,clip,width=\columnwidth]{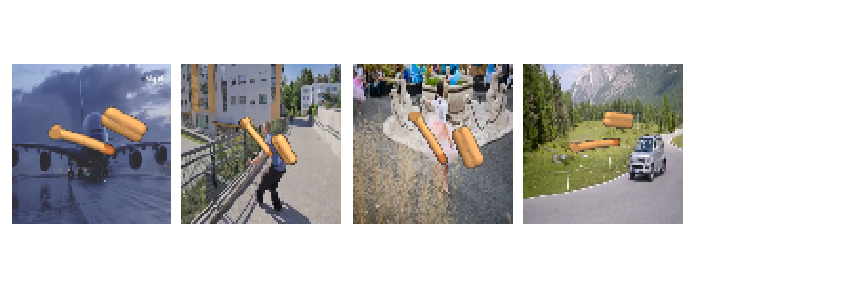}
\label{fig:subfigFS}}

\subfloat[][Reacher Easy]{
\includegraphics[trim=0 60 170 60,clip,width=\columnwidth]{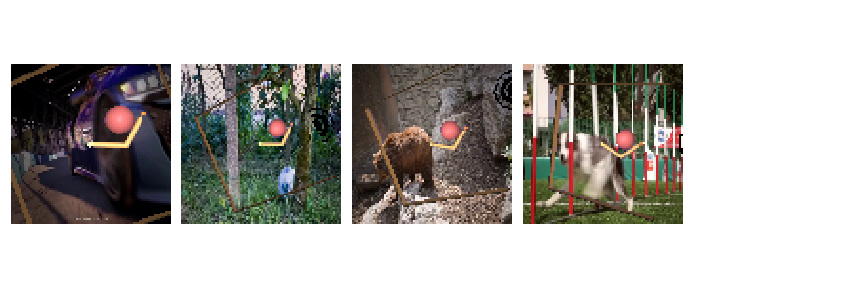}
\label{fig:subfigRE}}

\subfloat[][Walker Walk]{
\includegraphics[trim=0 60 170 60,clip,width=\columnwidth]{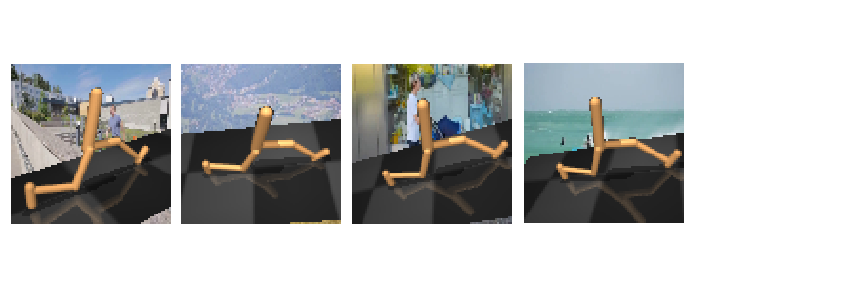}
\label{fig:subfigWW}}

\caption{\footnotesize DCS environments with distracting backgrounds and camera. The same semantic situation (agent position relative to environment) is shown from varying viewpoints and backgrounds. The trained agent needs to be robust to these perturbations.}
\label{fig:dcs_environment}
\end{center}
\end{figure}

In this section we benchmark our improvements in bisimulation learning on a challenging control benchmark with distractors. The goal is to evaluate the efficacy of the improved bisimulation metrics under a variety of environment conditions and agent capacity, and to evaluate if bisimulation provides additional benefits beyond those achieved with standard data augmentation. We use 6 environments in the Distracting Control Suite (DCS): Ball in cup catch (BiC), carpole swingup (CS), cheetah run (CR), finger spin (FS), reacher easy (RE), and walker walk (WW). Similarly to \cite{agarwal2021contrastive}, environments use dynamic background distractions from the DAVIS 2017 dataset \cite{pont20172017}, with $2$ videos used for training and $30$ for evaluation to test for out of distribution generalization. We also include an out-of-plane camera motion to further increase task difficulty, an example of the observations produced can be seen on Figure \ref{fig:dcs_environment}. Additional experiments on autonomous driving using CARLA \cite{dosovitskiy2017carla} are provided in Appendix \ref{sec:supplementary_carla}.

We test $6$ training setups: SAC as a baseline, DrQ, which adds low-level visual augmentation, DrQ $+$DBC, a reward-based bisimulation measure on top of DrQ, $\epsilon$-R, our entagled correction for reward-based bisimulation ( $G(\bm{z},\bm{a},\bm{z}',\bm{a}') = |R(\bm{z},\bm{a})-R(\bm{z}',\bm{a}')|$) on top of DrQ, DrQ$+$PSM, a policy-based bisimulation measure, and $\epsilon$-$\pi$, our entagled correction for policy-based bisimulation ( $G(\bm{z},\bm{a},\bm{z}',\bm{a}') = ||\mathbb{E}_{\pi(\bm{a}|\bm{z})}[\bm{a}]-\mathbb{E}_{\pi(\bm{a}'|\bm{z}')}[\bm{a}']||_1$) on top of DrQ. All methods share a common architecture and hyperparameters, described in Appendix \ref{sec:hyperparameters}; for bisimulation-based methods, we further tune the bisimulation strength hyperparameter $\beta$, only the results for the best performing hyperparameters are shown. 

\begin{figure}[h]
\centering
\subfloat{
\includegraphics[width=0.5\columnwidth]{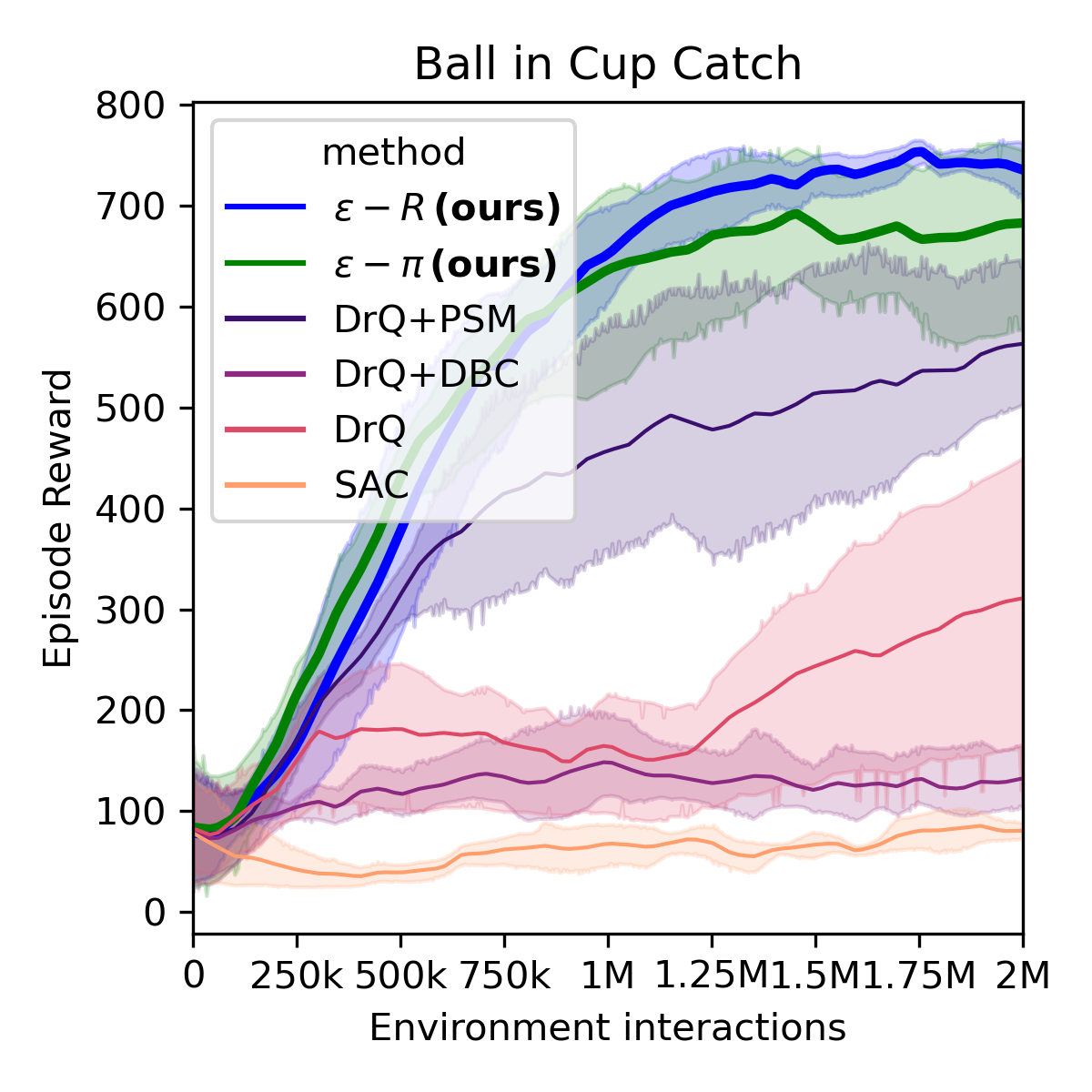}
\label{fig:subfig1}}
\subfloat{
\includegraphics[width=0.5\columnwidth]{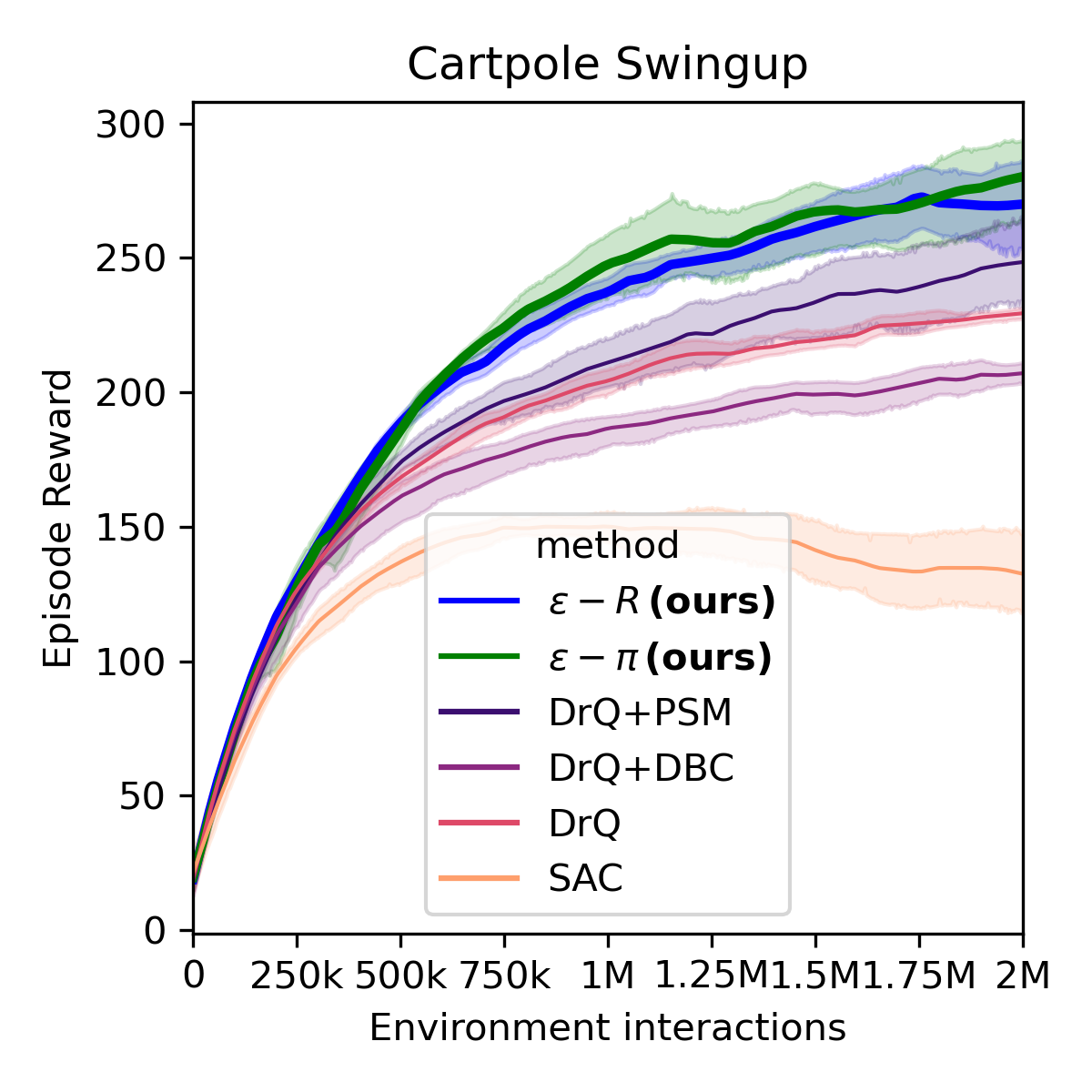}
\label{fig:subfig2}}

\vspace{.6em}
\subfloat{
\includegraphics[width=0.5\columnwidth]{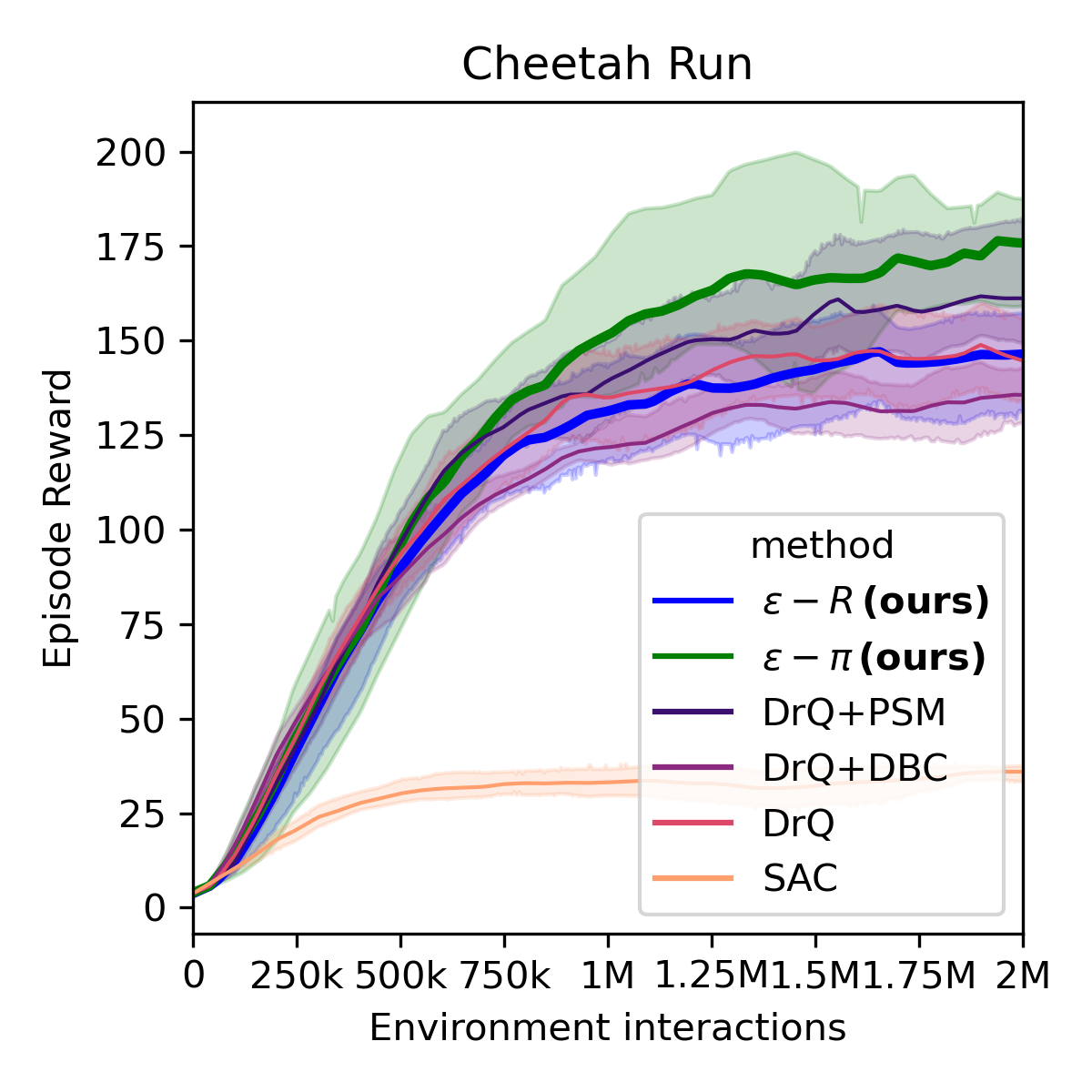}
\label{fig:subfig3}}
\subfloat{
\includegraphics[width=0.5\columnwidth]{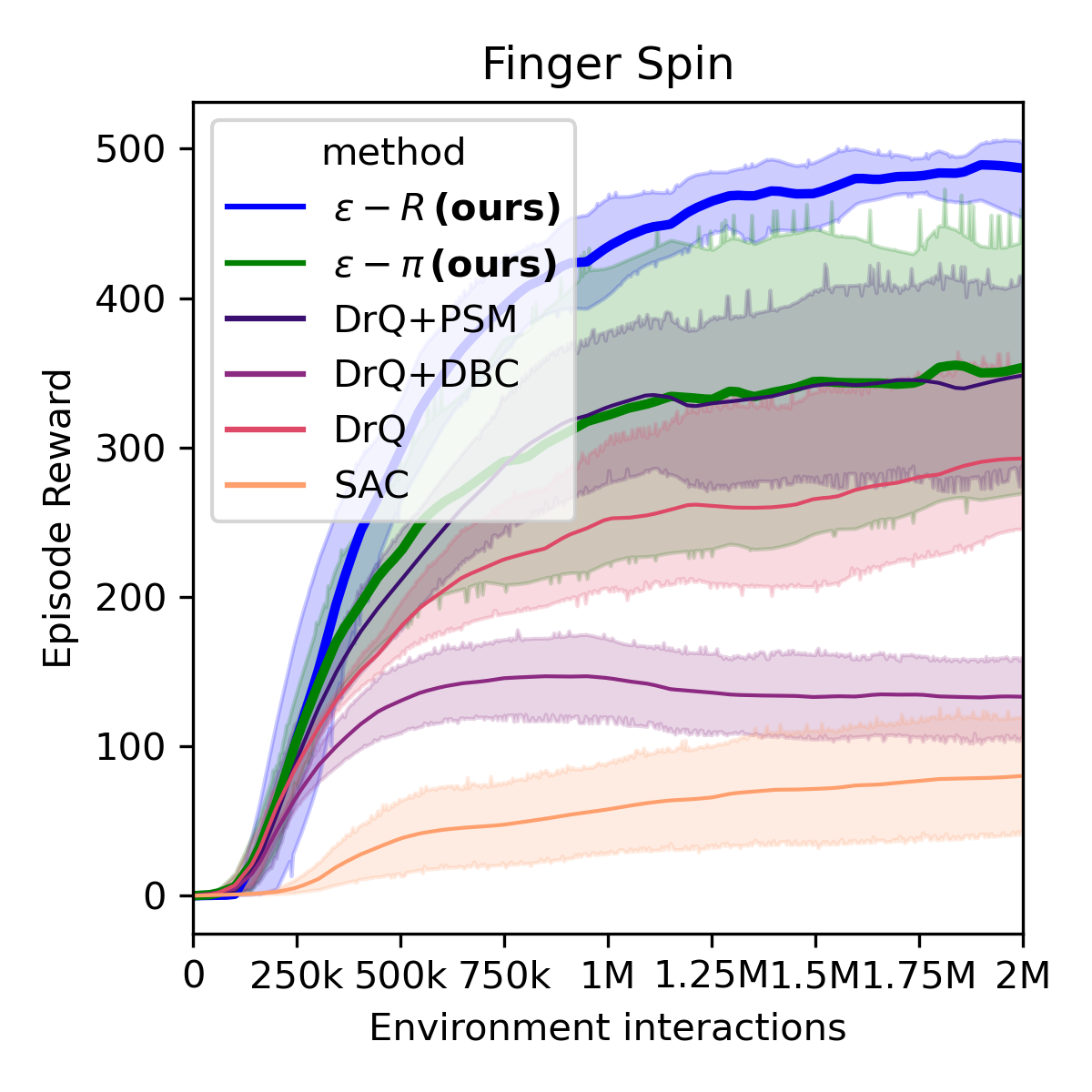}
\label{fig:subfig4}}

\vspace{.6em}
\subfloat{
\includegraphics[width=0.5\columnwidth]{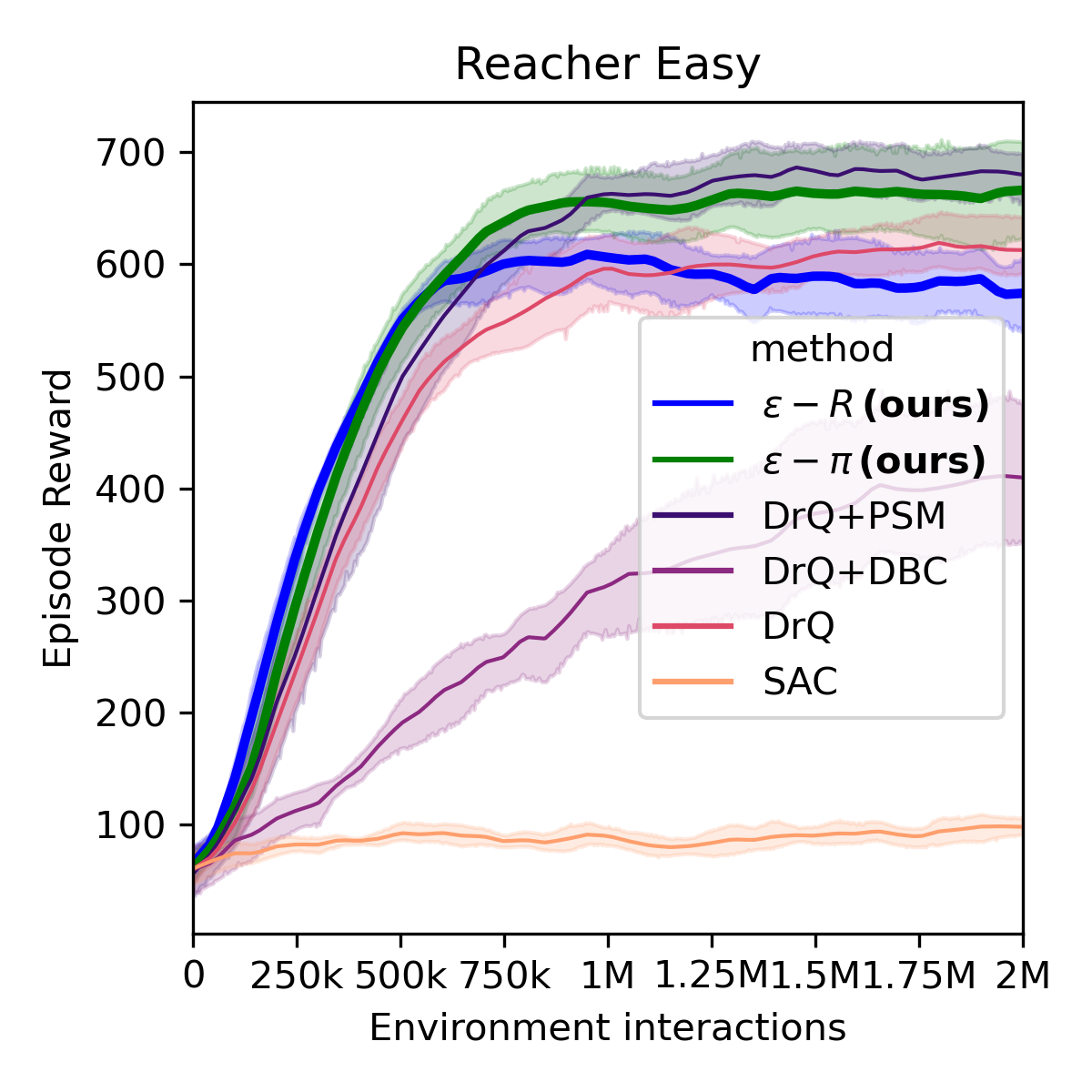}
\label{fig:subfig5}}
\subfloat{
\includegraphics[width=0.5\columnwidth]{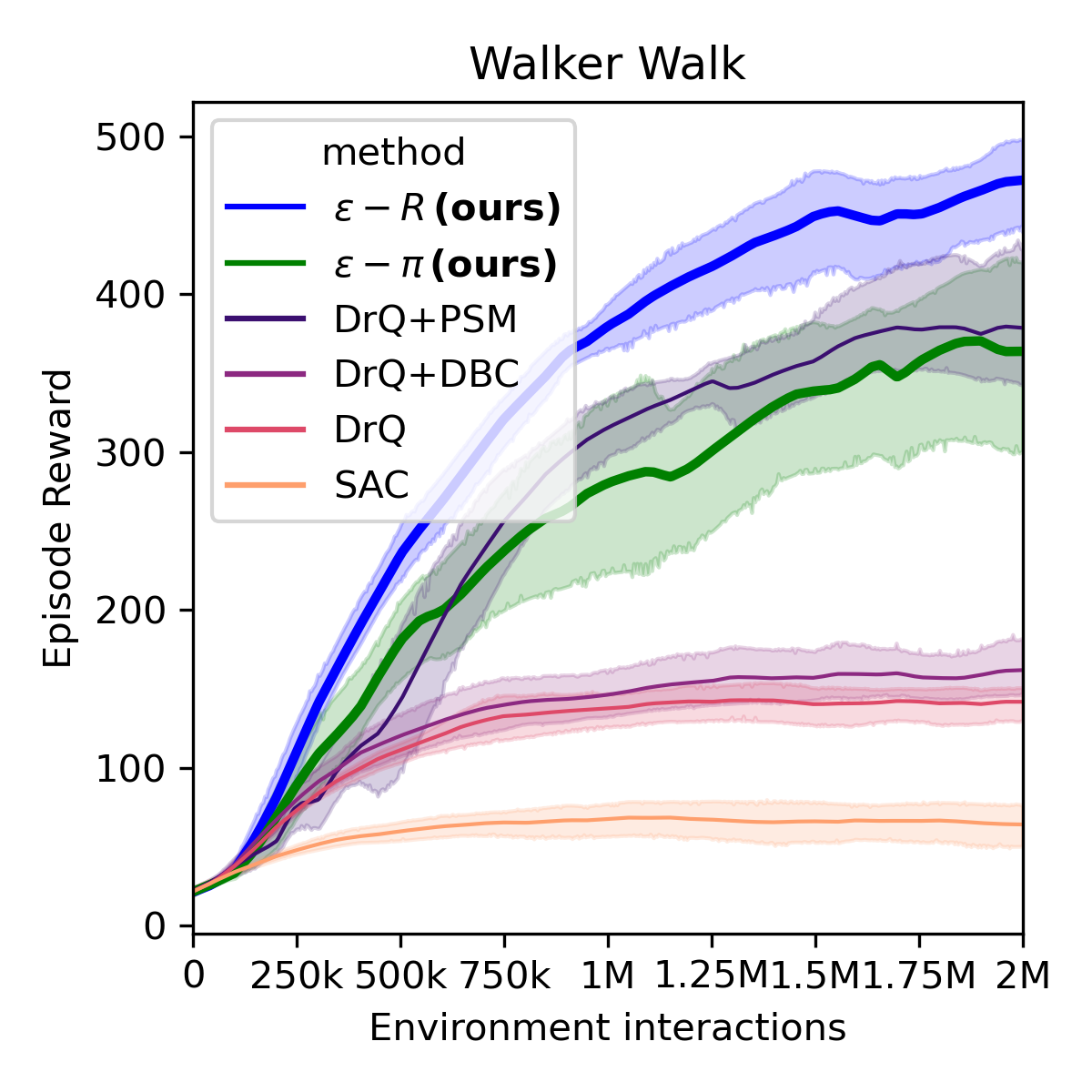}
\label{fig:subfig6}}
\caption{Episodic reward on test environments in DCS benchmark with hard camera distractions and $30$ background videos. Standard deviations are collected across $5$ independent seeds.}
\label{fig:dcs_results}
\end{figure}

\begin{figure}[h!]
\centering
\subfloat{
\includegraphics[width=\columnwidth]{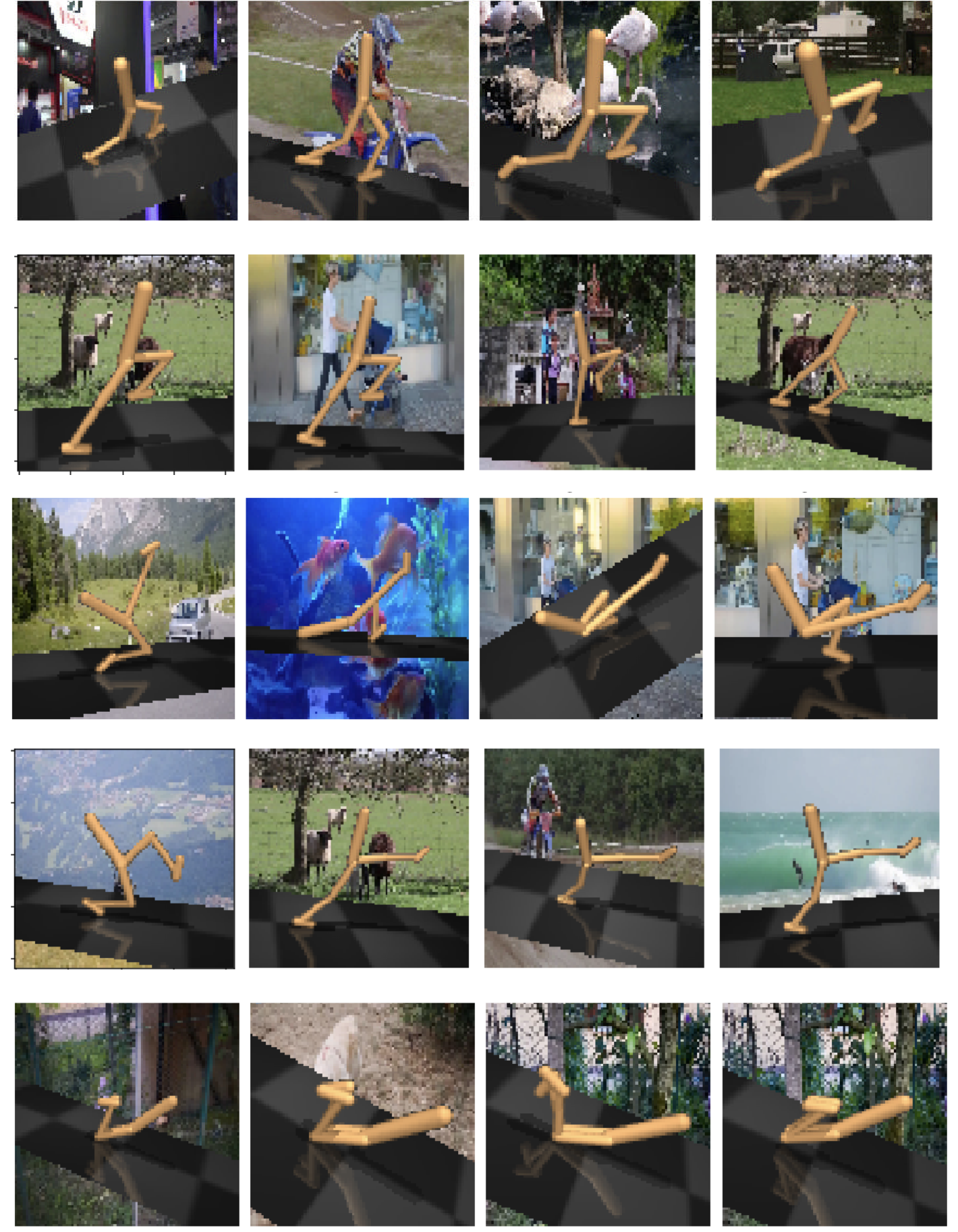}
}
\vspace{.5em}
\caption{Sample observations, alongside their closest bisimulation matches, acquired over 10 episodes on the Walker Walk environment with hard backgrounds and camera distractions. We acquired $5$ independent cameras, each with its own independent motion and distracting background, to provide novel viewpoints for the bisimulation comparisons. }
\label{fig:dcs_bisimulation}
\end{figure}

Table \ref{tab:base} shows the results with background distractions and camera movement, data augmentation in the form of DrQ improves on the SAC baseline, and bisimulation metrics improve on this even further, with $\epsilon$-$\pi$ and $\epsilon$-R providing the best overall results. We observe that the largest improvement provided by entanglement is seen on the DrQ+DBC to $\epsilon$-R comparison, in some cases boosting performance over $3\times$ over the unmodified algorithm; it also alleviates failiure cases as seen in BiC, FS, and RE, where DrQ+DBC is a net loss over just the DrQ baseline. Some environments (BiC, FS, WW) show significant improvement of the bisimulation methods w.r.t. data augmentation, though none are adversely affected by the inclusion of bisimulation. The training dynamics of these methods are shown in Figure \ref{fig:dcs_results}. Additionally, Figure \ref{fig:dcs_bisimulation} shows random observations, along with their closest bisimulation matches across other episodes and backgrounds on a small subset of 10 episodes acquired on the Walker Walk environment. Semantically similar situations are consistently identified regardless of background or camera position. 

\begin{table}[h]
    \caption{Episodic reward on Distracting Control Suite with DAVIS backgrounds and hard camera distractions, $2M$ environment steps. Mean and standard deviation computed across 5 seeds.}
    \label{tab:base}
    \centering
    \begin{adjustbox}{max width=\columnwidth}
    \begin{tabular}{c|cccccc}
    Method & BiC & CS & CR & FS & RE & WW\\
    \hline
     SAC & $ 240 \pm 9$ & $133 \pm	20$& $36	\pm 3$ & $80 \pm 51$ & $98 \pm	9$ & $64 \pm	17$\\
     DrQ & $ 311 \pm 181$ & $229 \pm 3$ & $145 \pm	14$ &$293 \pm	72$ & $613 \pm	33$ & $142	\pm 14$\\
     DrQ+DBC & $ 132 \pm 39$ & $207 \pm	5$ & $134	\pm 12$&  $144 \pm	32$ & $410 \pm	83$ & $150 \pm	21$\\
     $\bm{\epsilon}$\textbf{-R} & $ \bm{735 \pm 37} $& $270 \pm 22$ & $ 146 \pm	18$ & $\bm{487 \pm	33}$ & $574 \pm	42$& $\bm{472 \pm 37}$\\
     DrQ+PSE &$ 563 \pm 94 $ & $ 248 \pm	20$ & $161 \pm 23$ & $348 \pm	83 $& $\bm{680 \pm	29}$& $379 \pm	59$\\
     \textbf{$\bm{\epsilon}$-$\bm{\pi}$} & $683 \pm 103$ & $\bm{280 \pm 20}$ & $\bm{175 \pm	18}$ & $354 \pm	105$ & $666	\pm 57$ & $364 \pm	80$
    \end{tabular}
    \end{adjustbox}
\end{table}

\begin{figure}[h!]
\centering
\subfloat[][Ball in Cup Catch]{
\includegraphics[width=0.5\columnwidth]{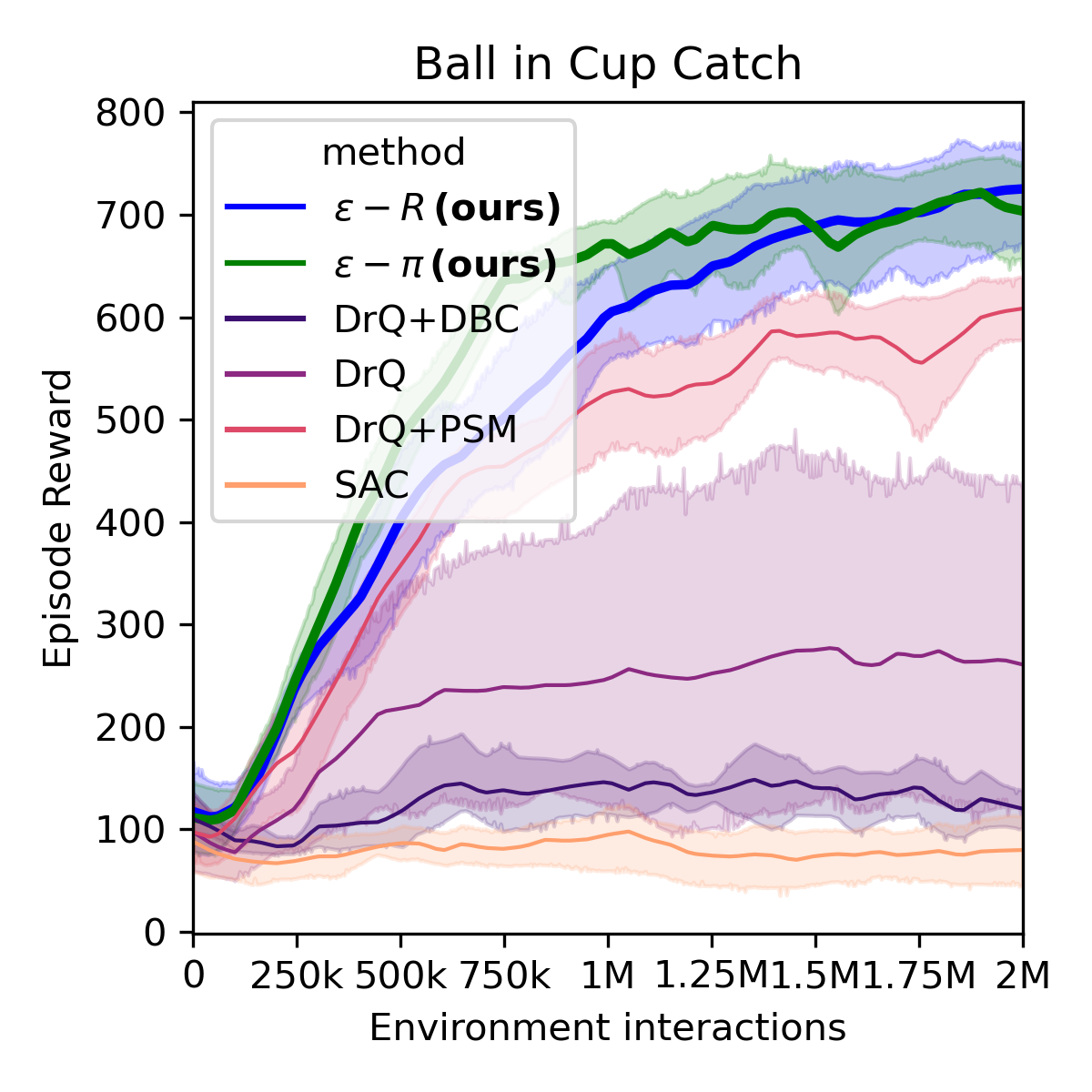}
\label{fig:lcsubfig1}}
\subfloat[][Cartpole Swingup]{
\includegraphics[width=0.5\columnwidth]{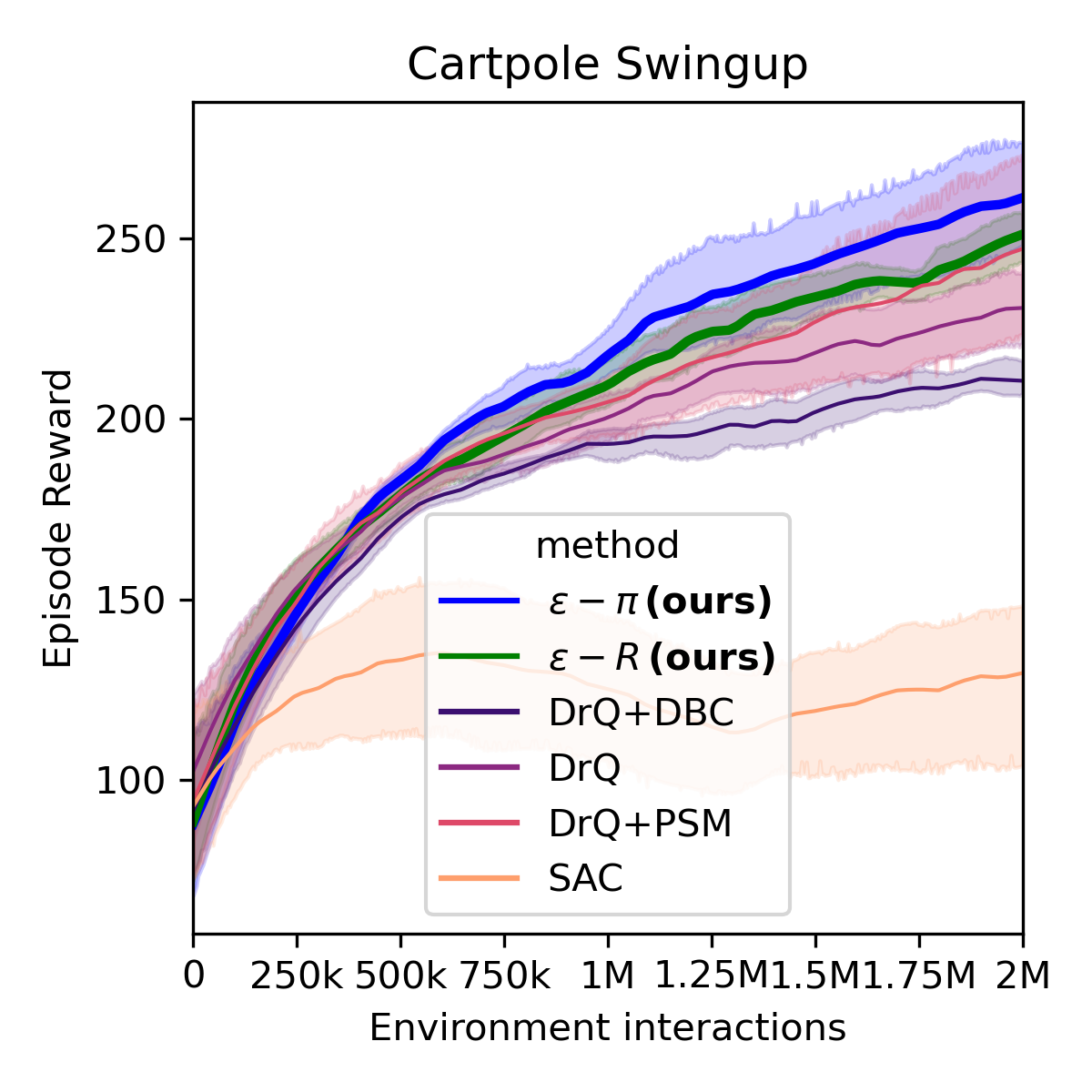}
\label{fig:lcsubfig2}}

\subfloat[][Cheetah Run]{
\includegraphics[width=0.5\columnwidth]{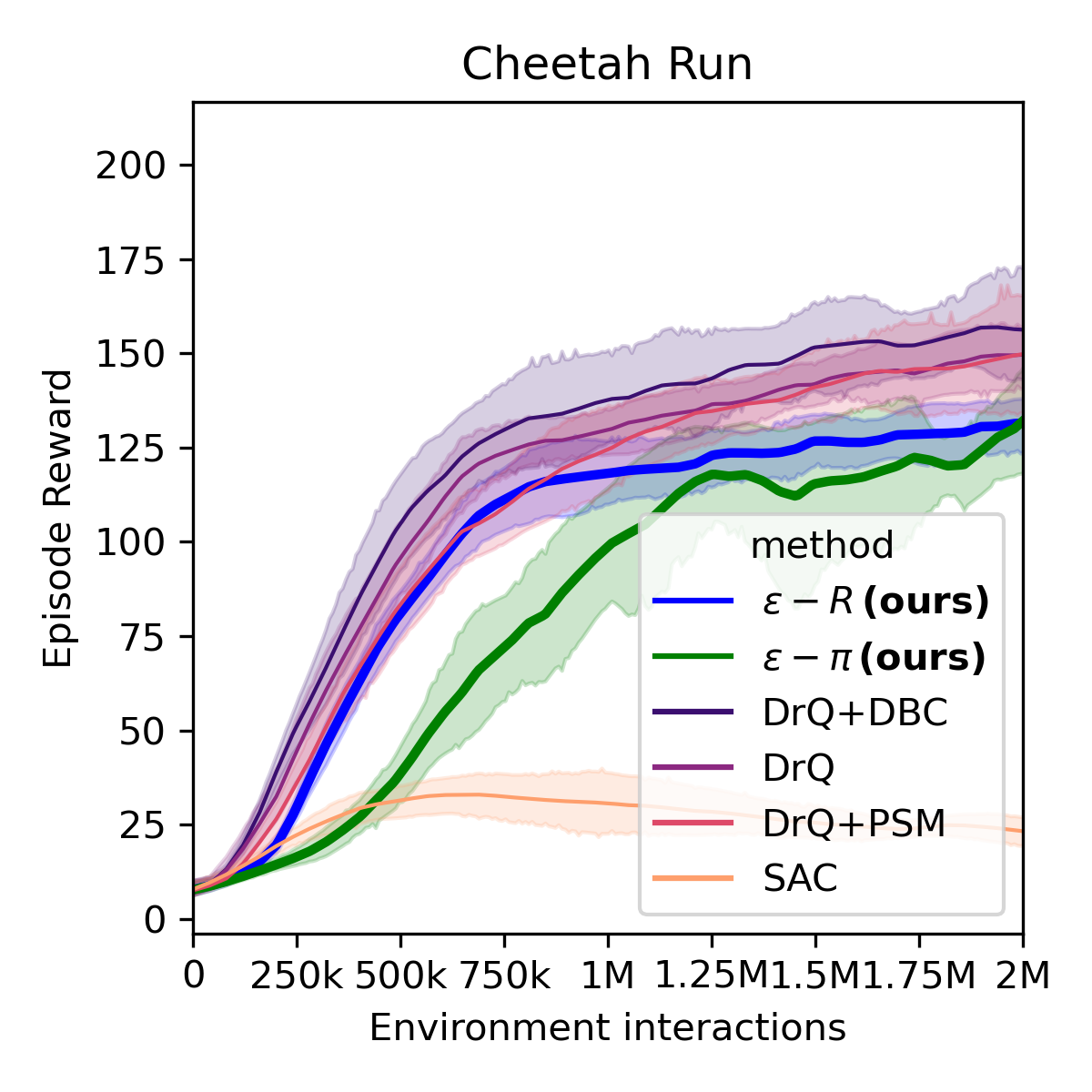}
\label{fig:lcsubfig3}}
\subfloat[][Finger Spin]{
\includegraphics[width=0.5\columnwidth]{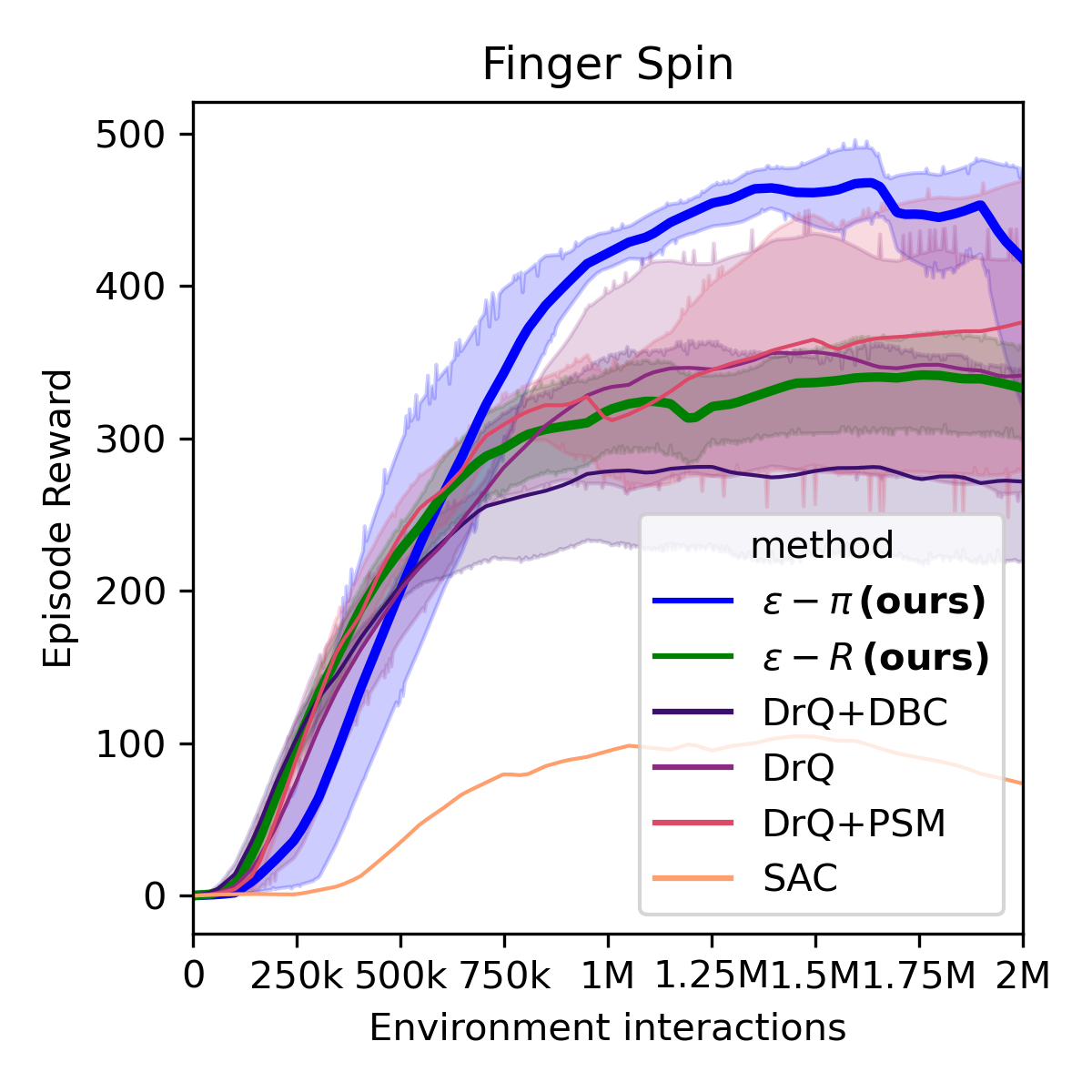}
\label{fig:lcsubfig4}}

\subfloat[][Reacher Easy]{
\includegraphics[width=0.5\columnwidth]{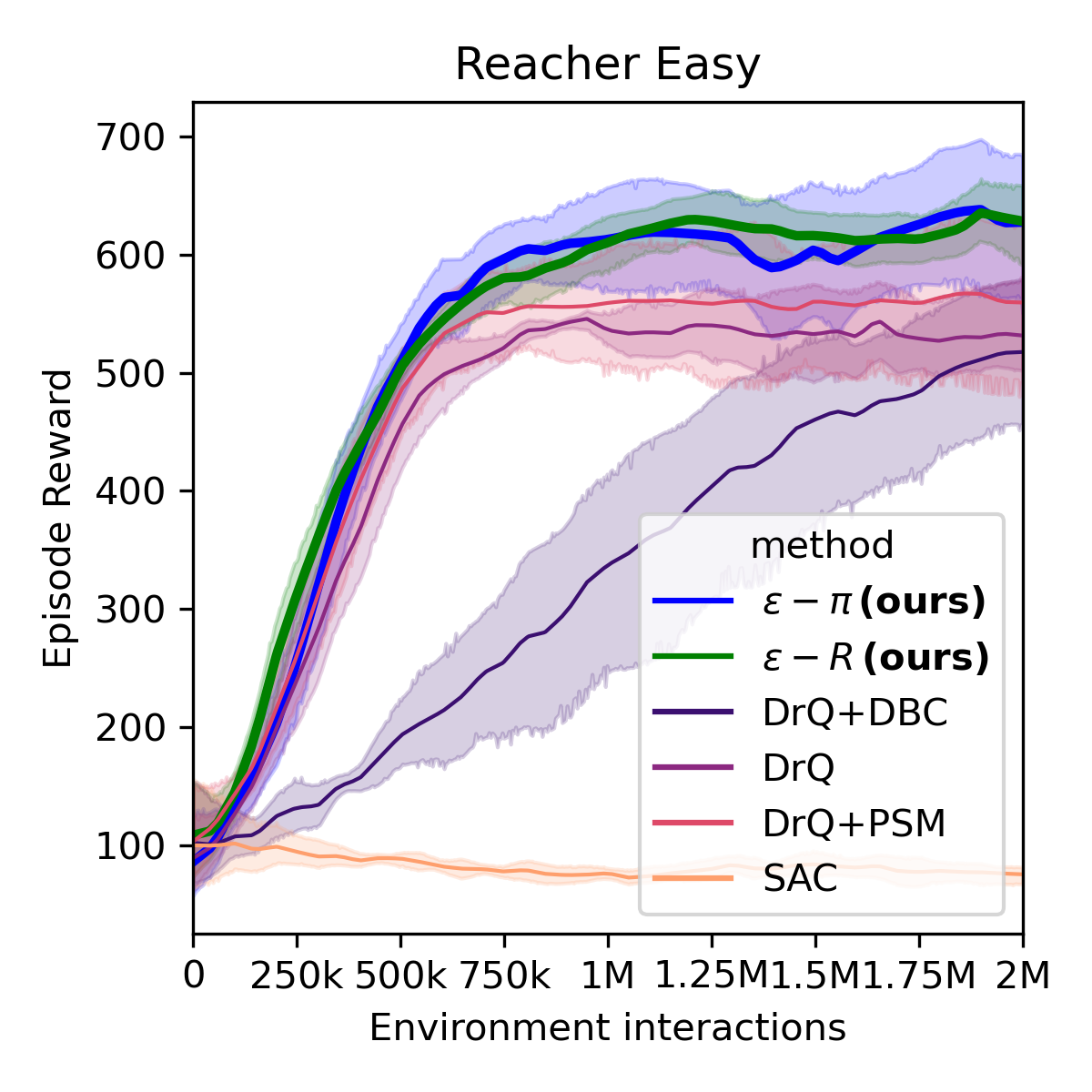}
\label{fig:lcsubfig5}}
\subfloat[][Walker Walk]{
\includegraphics[width=0.5\columnwidth]{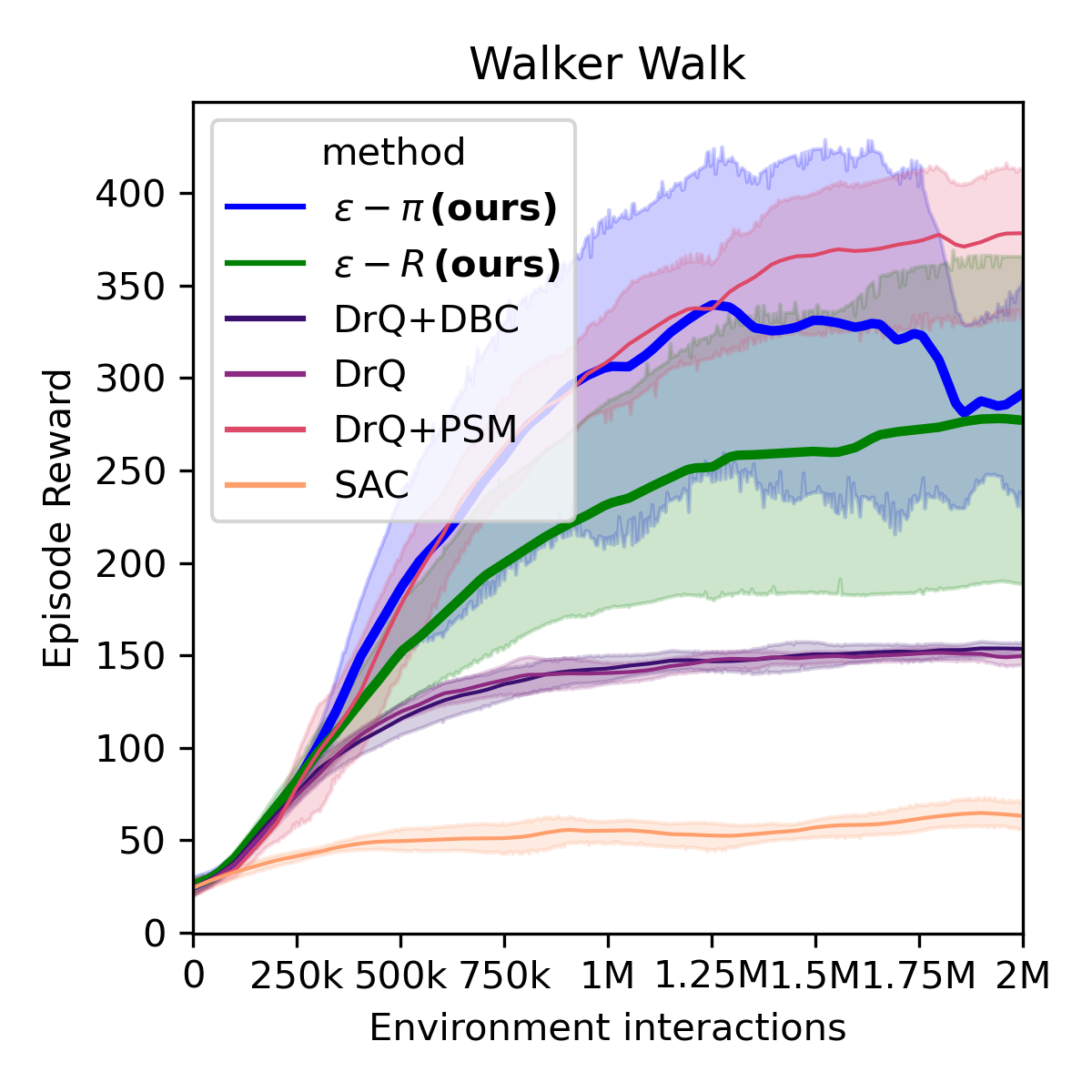}
\label{fig:lcsubfig6}}
\caption{Episodic reward on test environments in DCS benchmark with hard camera distractions and $30$ background videos from the DAVIS dataset. Low capacity architecture. Standard deviations are collected across $5$ independent seeds.}
\label{fig:dcs_results_lc}
\end{figure}

To test if network capacity plays a major role in these results, we repeat these experiments using a smaller architecture in Table \ref{tab:lowcap}. Here, all hidden layers have their size reduced from $1000$ to $200$ units. The results show a loss of performance across the board, suggesting that capacity is being properly utilized on the larger architecture size. Bisimulation-based methods still show the best overall performances, with reward-based bisimulation methods taking a comparatively larger hit from the reduced network capacity. The same environments that previously exhibited large performance differences between DrQ and the bisimulation-based methods continue to do so. Figure \ref{fig:dcs_results_lc} shows the training dynamics in the low-capacity scenario.

\begin{table}[h]
    \centering
    \caption{Episodic reward on Distracting Control Suite with DAVIS backgrounds and hard camera distractions, $2M$ environment steps, low capacity network. Mean and standard deviation computed across 5 seeds.}
    \label{tab:lowcap}
    \begin{adjustbox}{max width=\columnwidth}
    \begin{tabular}{c|cccccc}
    Method & BiC & CS & CR & FS & RE & WW\\
    \hline
     SAC & $78 \pm 43 $ & $130 \pm 28$ & $31 \pm 7$ & $74 \pm 35$ & $75 \pm 10$ & $63 \pm 10$\\
     DrQ & $261 \pm 206$ & $231 \pm 13$ & $156 \pm 19$ & $341 \pm 94$ & $532 \pm 52$ & $150 \pm 6$\\
     DrQ+DBC & $120 \pm 21$ & $ 211 \pm 6$ & $162 \pm 17$ & $276 \pm 89$ & $518 \pm 80$ & $147 \pm 8$\\
     $\bm{\epsilon}$\textbf{-R} & $\bm{725 \pm 62}$ & $251 \pm 9$ & $124 \pm 7$ & $333 \pm 37$ & $\bm{628 \pm 41}$ & $277 \pm 116$\\
     DrQ+PSE & $608 \pm 36$ & $247 \pm 33$ & $\bm{171 \pm 40}$ & $376 \pm 121$ & $560 \pm 90$ & $\bm{378 \pm 51}$\\
     \textbf{$\bm{\epsilon}$-$\bm{\pi}$} & $704 \pm 64$ & $\bm{261 \pm 18}$ & $117 \pm 52$ & $\bm{418 \pm 97}$ & $\bm{628 \pm 80}$ & $292 \pm 75$
    \end{tabular}
    \end{adjustbox}

\end{table}

\begin{table}[h]
    \centering
    \caption{Episodic reward on Distracting Control Suite with DAVIS backgrounds and no camera distractions, $2M$ environment steps. Mean and standard deviation computed across 5 seeds.}
    \label{tab:nocam}
    \begin{adjustbox}{max width=\columnwidth}
    \begin{tabular}{c|cccccc}
    Method & BiC & CS & CR & FS & RE & WW\\
    \hline
     $\bm{\epsilon}$\textbf{-R} & $771 \pm	9$ & $\bm{367 \pm	29}$ & ${239 \pm	48}$ & $419 \pm	129$ & $642 \pm	35$ & $\bm{541 \pm	77}$\\
     DrQ+PSE & $673 \pm	328$& $348 \pm	51$ & $\bm{305 \pm	33}$ & $\bm{573 \pm	131}$ & $\bm{703 \pm	11}$ & $411 \pm	138$\\
     \textbf{$\bm{\epsilon}$-$\bm{\pi}$} & $\bm{782 \pm	84}$ & $348  \pm	51$ & $249 \pm	58$ & $528 \pm	77$ & $696 \pm	29$ & $463 \pm		103$
    \end{tabular}
    \end{adjustbox}
\end{table}

Out of the two distractions present in our benchmark, camera movement is the one least addressed by low-level augmentation techniques like the random cropping added in DrQ. For this reason, we remove camera distractions and rerun the experiments, which are shown in Table \ref{tab:nocam}. Without camera distractions, all environments significantly improve their performance.

\section{Discussion}

This paper examines current methods for extracting task-relevant state in environments that present significant task-irrelevant distractions. We focus on bisimulation-based methods, their practical implementations, and some of the consistent estimation errors present in current methods for continuous action spaces. We propose \textit{entangled} bisimulation, a new bisimulation framework that is able to eliminate estimation bias for a broad class of user-specified invariances, and is trivial to implement in many practical scenarios. We enable the use of user-specified invariances throughout the representation learning process.  We empirically show how this improved estimation positively impacts performance on a variety of environments for the common reward- and policy-based invariances, and how bisimulation measures interact with standard data augmentation techniques.

Our experiments show that certain, but not all, environments are improved by the addition of bisimulation, even beyond standard data augmentation. No environments see their performance negatively affected by this addition, suggesting that this technique can be added into standard RL practice without adverse effects. We also observed that, with proper baseline architectures, reward- and policy-based  bisimulation yields similar results after correcting for estimation biases. 

Future work should be aimed at understanding why certain environments show large performance differences with bisimulation-based measures, while others remain unaffected. A proper understanding of what properties of an environment or a distraction are directly addressed by either data augmentation or bisimulation is needed. 

\bibliography{main}
\bibliographystyle{icml2022}
\clearpage

\appendix
\onecolumn
\color{black}
\section{Theorem Proofs}
Here we provide the proof for all theorems stated in this work.

\paragraph{Theorem \ref{theofixedpoint}.} Define the metric operator $\mathcal{F}^\epsilon: \mathbb{M} \rightarrow \mathbb{M}$ as 
\begin{equation}
    \mathcal{F}_\epsilon(d)(\bm{z},\bm{z'}) = \mathbb{E}_{\bm{a},\bm{a}'\sim \gamma^\epsilon(\pi(\cdot|\bm{z}),\pi(\cdot|\bm{z}'))} [G(\bm{z},\bm{a},\bm{z}',\bm{a}') + c \mathcal{W}_1(P(\bm{z}_{+}|\bm{z},\pi), P(\bm{z}'_{+}|\bm{z}',\pi); d)].
    \label{eq:epsilonbisimtheosupp}
\end{equation}
Then, with $c \in [0,1)$, $\mathcal{F}_\epsilon$ has at least a fixed point $d_\epsilon$ which is a bisimulation metric.

\paragraph{Proof.} We first observe that, since G is non-negative
\begin{equation}
\begin{array}{l}
          A\coloneqq \{(\bm{z}, \bm{z'}) : \mathbb{E}_{\bm{a},\bm{a}'\sim \gamma^\epsilon(\pi(\cdot|\bm{z}),\pi(\cdot|\bm{z}'))} G(\bm{z},\bm{a},\bm{z}',\bm{a}') = 0\},\\
          B \coloneqq \{(\bm{z}, \bm{z'}): G(\bm{z},\bm{a},\bm{z}',\bm{a}') =0, \forall  \bm{a},\bm{a}' \in \textit{Supp}(\gamma^\epsilon(\pi(\cdot|\bm{z}),\pi(\cdot|\bm{z}'))) \\
          (\bm{z}, \bm{z'}) \in A \iff (\bm{z}, \bm{z'}) \in B,
\end{array}
\end{equation}
\noindent except on a zero-measure set in $\gamma^\epsilon(\pi(\cdot|\bm{z}),\pi(\cdot|\bm{z}'))$.

The rest of the proof largely mimics the proof of Theorem 4.5 in \cite{ferns2004metrics}, and its adaptation in Theorem 2 in \cite{castro2020scalable}, which is Theorem 1 in this paper, we largely use the same notation when appropriate, and refer the reader to \cite{ferns2004metrics} for the details ommited in this proof. Lemma 4.1 in \cite{ferns2004metrics} holds for definition \ref{def:entangledbisimrelation}. We make use of the same distance ordering on $\mathbb{M}$ where $d\le d' \iff d(\bm{z},\bm{z'})\le d'(\bm{z},\bm{z'}), \, \forall \bm{z},\bm{z'} \in \mathcal{Z}^2$, this produces an $\omega$-cpo (an $\omega$ complete partial order) with bottom $\perp$ the everywhere zero distance. Since we also use $\mathcal{W}_1$ to measure state distribution distances in $\mathcal{F}^\epsilon$, then Lemma 4.4 in \cite{ferns2004metrics} also applies (continuity of $\mathcal{W}_1$). 

The remaining piece for the proof is showing that $\mathcal{F}^\epsilon$ is continuous, meaning that for every $\omega$-chain $\{d_n\}$ (an increasing sequence of metrics), then $\mathcal{F}^\epsilon(\cup_n \{d_n\}) = \cup_n \{\mathcal{F}^\epsilon(d_n)\}$, where $\cup_n \{d_n\}(\bm{z},\bm{z'}) = \sup_{n\in \mathbb{N}} d_n(\bm{z},\bm{z'})$. This follows from

\begin{equation}
    \begin{array}{rl}
        \mathcal{F}^\epsilon(\cup_n \{d_n\}) & =  \mathbb{E}_{\bm{a},\bm{a}'\sim \gamma^\epsilon(\pi(\cdot|\bm{z}),\pi(\cdot|\bm{z}'))} [G(\bm{z},\bm{a},\bm{z}',\bm{a}') \\&+ c \mathcal{W}_1(P(\bm{z}_{+}|\bm{z},\bm{a}), P(\bm{z}'_{+}|\bm{z}',\bm{a}'); \cup_n \{d_n\})],  \\
        & =  \sup_{n\in \mathbb{N}}\big\{\mathbb{E}_{\bm{a},\bm{a}'\sim \gamma^\epsilon(\pi(\cdot|\bm{z}),\pi(\cdot|\bm{z}'))} [G(\bm{z},\bm{a},\bm{z}',\bm{a}') \\&+ c  \mathcal{W}_1(P(\bm{z}_{+}|\bm{z},\bm{a}), P(\bm{z}'_{+}|\bm{z}',\bm{a}'); d_n)]\big\},  \\
        & =  \sup_{n\in \mathbb{N}}\mathcal{F}^\epsilon( d_n),  \\
    \end{array}
\end{equation}

Where the second equality follows by the continuity of $\mathcal{W}_1$ w.r.t. the metric $d_n$. The rest of the proof follows \cite{ferns2004metrics}.

\paragraph{Lemma \ref{lemma:otherdefinitions}.} Let $E^\pi$ be the largest $\pi$-bisimulation relation, and let $E^{PSM}$ likewise be the largest average-action bisimulation relation in \cite{agarwal2021contrastive}. The entangled bisimulation relation $E^\epsilon$ satisfies
\begin{equation}
\begin{array}{rl}
     E^\epsilon &\subseteq E^\pi  \\
     E^\epsilon &\subseteq E^{PSM}  \\
\end{array}
\end{equation}
for the choice of state-action similarity metrics $G(\bm{z},\bm{a},\bm{z}',\bm{a}') = ||R(\bm{z},\bm{a})-R(\bm{z}',\bm{a}')||_1$ and $G(\bm{z},\bm{a},\bm{z}',\bm{a}') =  ||\mathbb{E}_{\pi(\bm{a}|\bm{z})}[\bm{a}]-\mathbb{E}_{\pi(\bm{a}'|\bm{z}')}[\bm{a}']||_1$ respectively

\paragraph{Proof.} The proof is identical in both cases, so we focus on the reward-based case with $G(\bm{z},\bm{a},\bm{z}',\bm{a}') = ||R(\bm{z},\bm{a})-R(\bm{z}',\bm{a}')||_1$. From Jensen's inequality we have
\begin{equation}
\begin{array}{rl}
    \mathbb{E}_{{\bm{a}, \bm{a'}\sim \gamma^\epsilon(\pi(\cdot|\bm{z}),\pi(\cdot|\bm{z}'))}} [G(\bm{z},\bm{a},\bm{z}',\bm{a}')] 
    &\ge ||\mathbb{E}_{\bm{a}\sim \pi(\cdot|\bm{z})}R(\bm{z},\bm{a})- \mathbb{E}_{\bm{a}'\sim \pi(\cdot|\bm{z}')}R(\bm{z}',\bm{a}')||_1  \\
    &= ||R(\bm{z},\pi)- R(\bm{z}',\pi)||_1.  \\
\end{array}
\end{equation}

And, likewise, for any metric $d$ we have
\begin{equation}
\begin{array}{rl}
    \mathbb{E}_{{\bm{a}, \bm{a'}\sim \gamma^\epsilon(\pi(\cdot|\bm{z}),\pi(\cdot|\bm{z}'))}}\mathcal{W}_1(P(\bm{z}_{+}|\bm{z},\bm{a}), P(\bm{z}'_{+}|\bm{z}',\bm{a}'); d) &\ge\mathcal{W}_1(P(\bm{z}_{+}|\bm{z},\pi), P(\bm{z}'_{+}|\bm{z}',\pi); d)
\end{array}
\end{equation}

Then, we observe that $\mathcal{F}^\epsilon(d) \ge \mathcal{F}^\pi(d)$, and therefore $d^\epsilon \ge d^\pi$. From this we conclude that any state pair $(\bm{z}, \bm{z'}) \in E^\epsilon$ satisfies $0 \le d^\pi(\bm{z}, \bm{z'}) \le  d^\epsilon(\bm{z}, \bm{z'}) =0$ and therefore $(\bm{z}, \bm{z'}) \in E^\pi$.

\paragraph{Theorem \ref{theoMain}.} Given an MDP $\mathcal{M}$, policy $\pi$, and a state-action similarity metric $G$. Define the metric operator $\mathcal{F}_{\bar{\epsilon}}: \mathbb{M} \rightarrow \mathbb{M}$ as 
\begin{equation}
    \mathcal{F}_{\bar{\epsilon}}(d)(\bm{z},\bm{z'}) =\mathbb{E}_{\substack{(\bm{a}, \bm{a'})\sim\gamma^\epsilon(\pi(\cdot|\bm{z}),\pi(\cdot|\bm{z}'))\\(\bm{z}_+,\bm{z}'_+)\sim \gamma^\epsilon(P(\cdot|\bm{z},\bm{a}),P(\cdot|\bm{z}',\bm{a}'))
     }}[G(\bm{z},\bm{a},\bm{z}',\bm{a}') +c d_\phi(\bm{z}_{+}, \bm{z}'_{+})].
    \label{eq:epsilonuboperatorsupp}
\end{equation}

Then, $\mathcal{F}_{\bar{\epsilon}}$ has at least a fixed point $d_{\bar{\epsilon}}$ satisfying $d_{\bar{\epsilon}}\ge d_{\epsilon}$, and $ (\bm{z}, \bm{z'}) \in E^\epsilon \rightarrow d_{\bar{\epsilon}}(\bm{z}, \bm{z'})=0$, in particular, $d_{\bar{\epsilon}}(\bm{z}, \bm{z})=0, \, \forall \bm{z}\in \mathcal{Z}$.

Further, if the state transition function is coordinate independent ($P(\bm{z}_+\mid \bm{z},\bm{a}) =\Pi_i P_i(z_{i,+}\mid\bm{z},\bm{a})\, \forall \bm{z}_+,\bm{z},\bm{a} $), and $d_{\bar{\epsilon}}$ is of the form 

\begin{equation}
    d_{\bar{\epsilon}}(\bm{z},\bm{z}') = \sum\limits_{\substack{i=[n]\\j=[p]}}w_{i,j}|z_i-z'_i|^j, \; w_{i,j} \ge 0 \forall i,j, \; p> 0,
\end{equation}
then the bound is tight, that is
\begin{equation}
    \begin{array}{rl}
     d_\epsilon(\bm{z},\bm{z}') & = d_{\bar{\epsilon}}(\bm{z},\bm{z}')\\
     &=\mathbb{E}_{\substack{(\bm{a}, \bm{a'})\sim\gamma^\epsilon(\pi(\cdot|\bm{z}),\pi(\cdot|\bm{z}'))\\(\bm{z}_+,\bm{z}'_+)\sim \gamma^\epsilon(P(\cdot|\bm{z},\bm{a}),P(\cdot|\bm{z}',\bm{a}'))
     }}[G(\bm{z},\bm{a},\bm{z}',\bm{a}') +c d_\epsilon(\bm{z}_{+}, \bm{z}'_{+})].
    \end{array}
    \label{eqExpectedBisimTightsupp}
\end{equation}

\paragraph{Proof.} The proof will be split in three parts, we first prove that $d_{\bar{\epsilon}}$ exists and is an upper bound for $d_\epsilon$ on every state pair,  we then prove that $d_{\bar{\epsilon}}$ is zero on $\epsilon$-bisimilar states, and, lastly, we prove that the bound is tight for coordinate-independent transition functions if the distance function satisfies the prescribed functional form. To reduce notation clutter throughout the proof, we drop the state and action conditionals by noting $ P(\bm{z}_+) \coloneqq P(\cdot|\bm{z},\bm{a}) $, and, likewise, $Q(\bm{z}_+') \coloneqq P(\cdot|\bm{z}',\bm{a}')$. Additionally, we denote $\gamma^\epsilon_P \coloneqq \gamma^\epsilon(P(\cdot|\bm{z},\bm{a}),P(\cdot|\bm{z},\bm{a}'))$ and $\gamma^\epsilon_\pi \coloneqq \gamma^\epsilon(\pi(\cdot\mid \bm{z}),\pi(\cdot\mid \bm{z}'))$.

\paragraph{Upper bound:} To prove the existence of the fixed point metric $d_{\bar{\epsilon}}$, we again mimic the proof of Theorem 4.5 in \cite{ferns2004metrics}, only now we need to prove that the operation $T(d) \coloneqq \mathbb{E}_{\substack{(\bm{a}, \bm{a'})\sim\gamma^\epsilon_\pi\\(\bm{z}_+,\bm{z}'_+)\sim \gamma^\epsilon_P}}[ d(\bm{z}_{+}, \bm{z}'_{+})]$ is continuous (the analog of Lemma 4.4 in that work). 

This amounts to observing that, on an $\omega$-chain\footnote{Recall that an $\omega$-chain satisfies $d_i \le d_j$ for every $j>i$, and also $\cup_n \{d_n\}(\bm{z},\bm{z'}) = \sup_{n\in \mathbb{N}} d_n(\bm{z},\bm{z'})$, we also define $d_i \le d_j \iff d_i(\bm{z},\bm{z'}) \le d_j(\bm{z},\bm{z'}) \, \forall \,(\bm{z},\bm{z'}) \in \mathcal{Z}^2$} $\{d_n\}$,
\begin{equation}
\begin{array}{rl}
    T(\cup_n \{d_n\}) &= \mathbb{E}_{\substack{(\bm{a}, \bm{a'})\sim\gamma^\epsilon_\pi\\(\bm{z}_+,\bm{z}'_+)\sim \gamma^\epsilon_P}}[ \cup_n \{d_n\}(\bm{z}_{+}, \bm{z}'_{+})],\\
    &= \sup_{n\in \mathbb{N}}\mathbb{E}_{\substack{(\bm{a}, \bm{a'})\sim\gamma^\epsilon_\pi\\(\bm{z}_+,\bm{z}'_+)\sim \gamma^\epsilon_P}}[ d_n(\bm{z}_{+}, \bm{z}'_{+})]\\
    &= \sup_{n\in \mathbb{N}} T(d_n).\\
\end{array}
\end{equation}
The rest of the proof of existence of a fixed point needs to show continuity of $\mathcal{F}_{\bar{\epsilon}}$, which is nearly identical to the proof for Theorem \ref{theofixedpoint}.

Now, to prove that $d_\epsilon \le d_{\bar{\epsilon}}$, we recall that $d_\epsilon = \cup_{n\in \mathbb{N}} \{\mathcal{F}^n_{\epsilon}(\perp)\}$ where $\perp$ is the everywhere-zero metric and $\mathcal{F}^n_{\epsilon}(\perp)$ is the result of recursively applying $\mathcal{F}_{\epsilon}$ to $\perp$ n times; and, likewise $d_{\bar{\epsilon}} = \cup_{n\in \mathbb{N}} \{\mathcal{F}^n_{\bar{\epsilon}}(\perp)\}$. From this, it suffices to prove that, if $d \le d'$ then $\mathcal{F}_{\epsilon}(d) \le \mathcal{F}_{\bar{\epsilon}}(d')$, since this would imply that $\mathcal{F}^n_{\epsilon}(\perp) \le \mathcal{F}^n_{\bar{\epsilon}}(\perp), \forall\, n \in \mathbb{N}$. 

\begin{equation}
    \begin{array}{rl}
    \mathcal{F}_{\epsilon}(d)(\bm{z},\bm{z'}) &=  \mathbb{E}_{\bm{a},\bm{a}'\sim \gamma^\epsilon_\pi} [G(\bm{z},\bm{a},\bm{z}',\bm{a}') + c \mathcal{W}_1(P(\bm{z}_{+}|\bm{z},\pi), P(\bm{z}'_{+}|\bm{z}',\pi); d)], \\
    &\le  \mathbb{E}_{\bm{a},\bm{a}'\sim \gamma^\epsilon_\pi} [G(\bm{z},\bm{a},\bm{z}',\bm{a}') + c \mathcal{W}_1(P(\bm{z}_{+}|\bm{z},\pi), P(\bm{z}'_{+}|\bm{z}',\pi); d')], \\
    &\le  \mathbb{E}_{\bm{a},\bm{a}'\sim \gamma^\epsilon_\pi} [G(\bm{z},\bm{a},\bm{z}',\bm{a}') + c \mathbb{E}_{\bm{z_+},\bm{z_+}'\sim \gamma^\epsilon_P}[d'(\bm{z_+},\bm{z_+}')]], \\
    &=  \mathbb{E}_{\substack{\bm{a},\bm{a}'\sim \gamma^\epsilon_\pi\\\bm{z_+},\bm{z_+}'\sim \gamma^\epsilon_P}} [G(\bm{z},\bm{a},\bm{z}',\bm{a}') + c d'(\bm{z_+},\bm{z_+}')], \\
    &=  \mathcal{F}_{\bar{\epsilon}}(d)(\bm{z},\bm{z'}). \\
    \end{array}
\end{equation}

Here the first inequality derives from $d\le d'$ and the continuity of $\mathcal{W}_1$, and the second inequality stems from $\gamma^\epsilon_P\in \Gamma(P,Q)$ and the definition of $\mathcal{W}_1$.

\paragraph{$\epsilon$-similar states.} To prove that, if $\bm{z}, \bm{z}' \in E^\epsilon \rightarrow d_{\bar{\epsilon}}(\bm{z}, \bm{z}') = 0$, we again use that $d_\epsilon = \cup_{n\in \mathbb{N}} \{\mathcal{F}^n_{\epsilon}(\perp)\}$. Inductively, we have that $\mathcal{F}^0_{\epsilon}(\perp)(\bm{z}, \bm{z}')=0\,\forall \bm{z}, \bm{z}' \in E^\epsilon$(in fact, this is true everywhere), we now show that if $\mathcal{F}^j_{\epsilon}(\perp)(\bm{z}, \bm{z}')=0,\forall \bm{z}, \bm{z}' \in E^\epsilon $, then $\mathcal{F}^{j+1}_{\epsilon}(\perp)(\bm{z}, \bm{z}')=0,\forall \bm{z}, \bm{z}' \in E^\epsilon$. It suffices to observe that, under these conditions, if $\bm{z}, \bm{z}' \in E^\epsilon $ and $d : d(\bm{z}, \bm{z}')=0  \forall \bm{z}, \bm{z}' \in E^\epsilon $ then
\begin{equation}
\begin{array}{rl}
    \mathcal{F}_{\bar{\epsilon}}(d)(\bm{z},\bm{z}') =& \mathbb{E}_{\substack{\bm{a},\bm{a}'\sim \gamma^\epsilon_\pi\\\bm{z_+},\bm{z_+}'\sim \gamma^\epsilon_P}}[G(\bm{z},\bm{a},\bm{z}',\bm{a}') +c d(\bm{z}_{+}, \bm{z}'_{+})],\\
     &=0.\\
\end{array}
\end{equation}

\paragraph{Optimality of upper bound under conditions:} Finally, let $\bm{z}_{\setminus i} = \{z_j\}_{j\neq i}$. If $P(\bm{z}_+\mid\bm{z},\bm{a}) = \Pi_i P_i(z_{i,+}\mid\bm{z},\bm{a})\, \forall \bm{z}_+,\bm{z},\bm{a}$, and we further restrict the function class $d(\bm{z},\bm{z}')$ to be coordinate-wise separable and convex, we can first show that expected distances over couplings have a coordinate-factorized structure, that is, for any coupling $\gamma \in \Gamma(P,Q)$, we have

\begin{equation}
    \begin{array}{rl}
    \mathop{\mathbb{E}}\limits_{\bm{z},\bm{z}'\sim \gamma} [d(\bm{z},\bm{z}')] & =\int d(\bm{z},\bm{z}')^p \gamma(\bm{z},\bm{z}') d\bm{z}d\bm{z}'   \\
    & =\sum\limits_{i=1}^n \int d_i(z_i,z'_i) \gamma(\bm{z},\bm{z}') d\bm{z}d\bm{z}' \\
    & =\sum\limits_{i=1}^n \int d_i(z_i,z'_i) \gamma_i(z_i,z'_i)\gamma(\bm{z}_{\setminus i},\bm{z}'_{\setminus i}|z_i,z'_i) d\bm{z}d\bm{z}'\\
    & =\sum\limits_{i=1}^n \int_{z_i,z'_i} d_i(z_i,z'_i) \gamma_i(z_i,z'_i)\int_{\bm{z}_{\setminus i},\bm{z}'_{\setminus i}}\gamma(\bm{z}_{\setminus i},\bm{z}'_{\setminus i}|z_i,z'_i)d\bm{z}_{\setminus i} d\bm{z}'_{\setminus i} dz_idz'_i\\
    & =\sum\limits_{i=1}^n \int d_i(z_i,z'_i) \gamma(z_i,z'_i) dz_idz'_i\\
    & =\sum\limits_{i=1}^n \int d_i(z_i,z'_i) \gamma(z_i,z'_i)\mathop{\Pi}\limits_{j\in [n]\setminus i} \gamma(\hat{z}_j,\hat{z}'_j) d\hat{\bm{z}}_{\setminus i} d\hat{\bm{z}}_{\setminus i} dz_idz'_i\\
    & =\int d(\hat{\bm{z}},\hat{\bm{z}}') \mathop{\Pi}\limits_{i=1}^n \gamma(\hat{\bm{z}}_i,\hat{\bm{z}}'_i) d\hat{\bm{z}}d\hat{\bm{z}}' \\
    & =E_{\hat{\bm{z}},\hat{\bm{z}}'\sim \mathop{\Pi}\limits_{i=1}^n\gamma_i} [d(\hat{\bm{z}},\hat{\bm{z}}')] \\
    \end{array}
\end{equation}

In essence, the expected value of the coupling is the same as the one for the coordinate-independent coupling. The preceding result only requires $d$ to satisfy the coordinate independent assumption, also recall $\mathcal{W}_1(P,Q;d) =\inf_{\gamma \in \Gamma(P,Q)}\mathop{\mathbb{E}}\limits_{\bm{z},\bm{z}'\sim \gamma} [d(\bm{z},\bm{z}')]$.

A well-known formula the p-Wasserstein distance between univariate variables is
\begin{equation}W_p(P_i,Q_i)^p =  \int_{0}^{1} |F_i^{-1}(u)-G_i^{-1}(u)|^p du.\end{equation}

Where $F_i^{-1}(u)$ and $G_i^{-1}(u)$ are the inverse CDF's of the i-th coordinate of distributions $P(X)$ and $Q(Y)$ respectively. 

From the closed form solution, we deduce that $\gamma^\epsilon(P,Q)$ is the Wasserstein optimal coupling for any distance function of the form $d(\bm{z},\bm{z}')=\sum\limits_{\substack{i=[n]\\j=[p]}}w_{i,j}|z_i-z'_i|^j, \; w_{i,j} \ge 0 \forall i,j, \; p> 0$ if the transition distribution satisfies coordinate independence.

\clearpage
\section{Algorithm description}
\label{sec:algorithm_supplementary}

Here we provide a full description of the training algorithm, where an inverse dynamics loss is added to the standard SAC objective, along our proposed $\epsilon-$bisimulation loss.

\begin{algorithm}[h!]
\caption{Entangled SAC Algorithm}\label{alg:main}
\begin{algorithmic}
\FOR {Time $t=0$ to $\infty$}
\STATE Sample action: $\bm{a}_t \sim \pi_\theta(h_\theta(o_t))$
\STATE Get environment transition: $o_{t+1}, r_{t+1} \sim \text{Env}(\bm{a}_t)$
\STATE Store transition: $D \leftarrow D \cup \{o_t,\bm{a}_t, r_{t+1}, o_{t+1}\}$
\STATE Sample batch: $B\sim D$ 
\STATE Dynamics loss: $J_P =\frac{1}{n}\sum_{o_t, o_{t+1},\bm{a}_t \in B}[\ln P_\theta(\overline{h_
\theta(o_{t+1})}|\bm{a}_t, h_\theta(o_{t}))]$
\STATE Inverse dynamics loss: $J_{ID} = \frac{1}{n}\sum_{o_t, o_{t+1},\bm{a}_t \in B}||\bm{a}_t - A_\theta(h_\theta(o_t),h_\theta(o_{t+1}))||_2^2$
\STATE Reward loss: $J_R = \frac{1}{n}\sum_{o_t, r_{t},\bm{a}_t \in B}[||R_\theta(\bm{z}\sim P_\theta(\bm{z}|\bm{a}_t, h_\theta(o_{t})))-r_t||^2_2]$
\STATE Policy loss: $J_\pi$  \COMMENT{SAC policy update, Algorithm \ref{alg:sacpi}}
\STATE entangled bisimulation loss: $J_\epsilon$ \COMMENT{Entangled update, Algorithm \ref{alg:coreBisimsupp}}
\STATE Update parameters: $\theta \leftarrow \theta + \eta \nabla_\theta (J_P+J_{ID}+J_\pi+J_R +J_\pi + \lambda J_\epsilon)$ 
\ENDFOR
\end{algorithmic}
\end{algorithm}

\begin{algorithm}[h!]
\caption{Policy Algorithm}\label{alg:sacpi}
\begin{algorithmic}
\REQUIRE data batch $B=\{o_t,o_{t+1},\bm{a}_t, r_{t+1}\}_{i=1}^n$, policy $\pi_\theta$, critics $Q^j, \,j=\{1,2\}$, target critics $\hat{Q}^j, \,j=\{1,2\}$, observation encoder $h_\theta$.
\STATE Get future state value: $\overline{V} = \min\limits_{j=1,2}\hat{Q}^j_\theta(h_\theta(\bm{o_{t+1}}),\hat{\bm{a}}\sim\pi(\hat{\bm{a}}|h_\theta(\bm{o_{t+1}})))$
\STATE  Critic loss: $J_C =\frac{1}{n}\sum_{o_t, r_{t},\bm{a}_t, o_{t+1} \in B}\sum\limits_{j=1,2}||Q^j_\theta(h_\theta(\bm{o_{t}}),\bm{a}_{t}) - r_{t+1} - \gamma \overline{V}||_2^2$
\STATE Actor loss: $J_a =\frac{1}{n}\sum_{o_t,\bm{a}_t \in B}\ \overline{\alpha} \log \pi(\bm{a}_t|h_\theta(\bm{o_{t}})) - \min\limits_{j=1,2}\overline{Q}^j_\theta(h_\theta(\bm{o_{t}}),\bm{a}_{t})$
\STATE Train alpha: $J_\alpha = \alpha \log \overline{\pi(\bm{a}_t|h_\theta(\bm{o_{t}}))} - \mathcal{H}(\bm{a}_t|h_\theta(\bm{o_{t}}))$
\STATE Update target critics: $\hat{Q}^j_\theta = \tau {Q}^j_\theta + (1-\tau) \hat{Q}^j_\theta$
\STATE \OUTPUT Policy loss $J_\pi = J_a + J_C$
\end{algorithmic}
\end{algorithm}

\begin{algorithm}[h!]
\caption{Bisimulation Algorithm}\label{alg:coreBisimsupp}
\begin{algorithmic}
\REQUIRE Latent state batch $B=\{\bm{z}\}_{i=1}^n$, policy $\pi_\theta$, latent transition model $P_\theta(\cdot\mid,\bm{z},\bm{a})$, bisimulation distance function $d_\theta(\cdot,\cdot)$, similarity pseudometric $G$
\STATE Permute states $B'=\{\bm{z}'\} = \text{Perm}(\{\bm{z}\})$
\STATE Sample noise variables $\epsilon^A \sim \mathcal{N}(0, I^{|A|}), \epsilon^Z \sim \mathcal{N}(0, I^{|Z|})$
\STATE Compute tied actions $\bm{a} = \pi_\theta(\cdot|\bm{z}; \epsilon^A),\;  \bm{a}'=\pi_\theta(\cdot|\bm{z}'; \epsilon^A)$
\STATE Compute tied latent transitions $\bm{z}_+ = P_\theta(\cdot|\bm{z}, \bm{a};\epsilon^Z),\;  \bm{z}'_+=P_\theta(\cdot|\bm{z}', \bm{a}';\epsilon^Z)$
\STATE Bisimulation target $\hat{d}(\bm{z},\bm{z}') = G(\bm{z},\bm{a},\bm{z}',\bm{a}') +c d_\theta(\bm{z}_+,\bm{z}'_+)$
\STATE \OUTPUT bisimulation loss $J_\epsilon(\bm{z},\bm{z}') =\frac{1}{n}\sum\limits_{\bm{z'},\bm{z} \in, B,B'} ||d_\theta(\bm{z},\bm{z}') - \overline{\hat{d}(\bm{z},\bm{z}')}||^2_2$
\end{algorithmic}
\end{algorithm}

\section{Network Architecture and Hyperparameters}
\label{sec:hyperparameters}

We use the a similar architecture to \cite{zhang2020learning,agarwal2021contrastive}, with some key modifications to the encoder architecture. The encoder is implemented using a convnet trunk with $4$ layers and ReLU activations, each kernel has size $3$ with $32$ channels, stride $2$ on the first layer, and stride $1$ everywhere else; the output of the convnet trunk is projected into a $50$-dimensional vector with a fully-connected layer. Unlike \cite{zhang2020learning,agarwal2021contrastive}, the encoder is fully shared between the actor and the critic. Also unlike \cite{zhang2020learning,agarwal2021contrastive}, the encoder is applied to each frame in the input framestack independently, and the embeddings are then concatenated, yielding a $F\times 50$ embedding on a framestack with $F$ repeats ($F=3$ for all DCS environments). The independent embedding of each frame in the framestack is used to improve robustness on the distracting camera movements, since these distractions break low-level pixel correspondence in the same framestack.

The actor, critic, dynamics, inverse dynamics, and reward models are all implemented with MLPs with two hidden layers with $1024$ neurons each ($200$ for low capacity experiments) and ReLU activations. The continuous actions of the actor model are sampled via the reparametrization trick with a $\tanh$ nonlinearity, and the latent state prediction is parametrized as a  coordinate-independent Gaussian distribution.

Table \ref{tab:hyperparameters} summarizes the hyperparameters used for the RL experiments; Table \ref{tab:hyperparameters_dcs} summarizes the action repeats used for each DCS environment.

\begin{table}[]
    \centering
    \begin{tabular}{c|c}
    \toprule
        Parameter Name & Value  \\
        Replay buffer capacity & $100000$ \\
        Batch size & $512$ \\
        Discount $\gamma$ & $0.99$\\
        Optimizer & Adam \\
        Learning rate & $10^{-3}$ \\
        Critic target update frequency & $2$ \\
        Critic Q-function soft update rate $\tau$ & $0.01$\\
        Actor update frequency & $2$ \\
        Actor log stddev bounds & $[-10, 2]$ \\
        Temperature learning rate & $10^{-4}$ \\
        Initial Temperature & $0.1$ \\
        Entropy Target & $-\text{num actions}$\\
        Image augmentation pad (DrQ)& $4$\\
        
    \end{tabular}
    \caption{Architecture and training hyperparameters}
    \label{tab:hyperparameters}
\end{table}

\begin{table}[]
    \centering
    \begin{tabular}{c|c}
    \toprule
        Environment & action repeats  \\
        Ball in cup Catch & 4\\
        Cartpole Swingup & 8\\
        Cheetah Run & 4\\
        Finger Spin & 2\\
        Walker Walk & 2\\
        Reacher Easy & 4\\
    \end{tabular}
    \caption{Action repeat parameter used in DCS environments}
    \label{tab:hyperparameters_dcs}
\end{table}

\clearpage

\section{Autonomous Driving Experiments}
\label{sec:supplementary_carla}

\begin{figure}[h]
\begin{center}
\footnotesize
\setlength{\tabcolsep}{0pt}

\subfloat[][]{
\includegraphics[width=0.45\columnwidth]{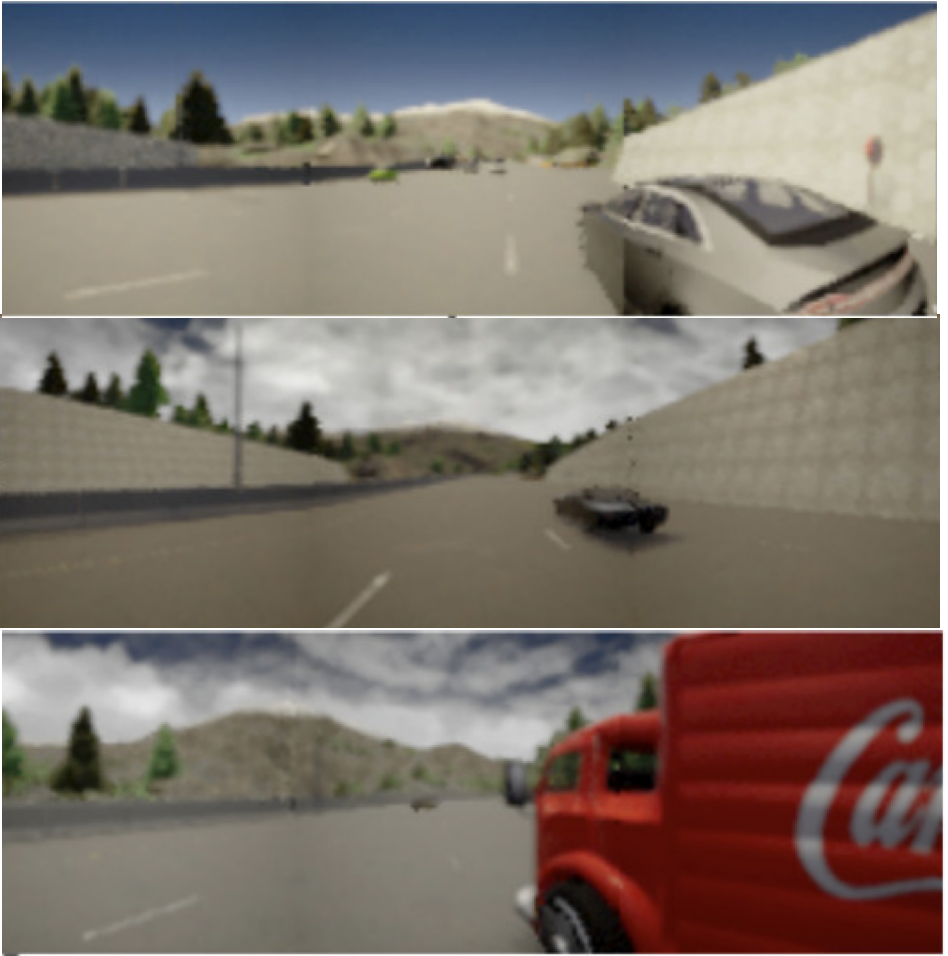}}
\subfloat[][]{
\includegraphics[width=0.45\columnwidth]{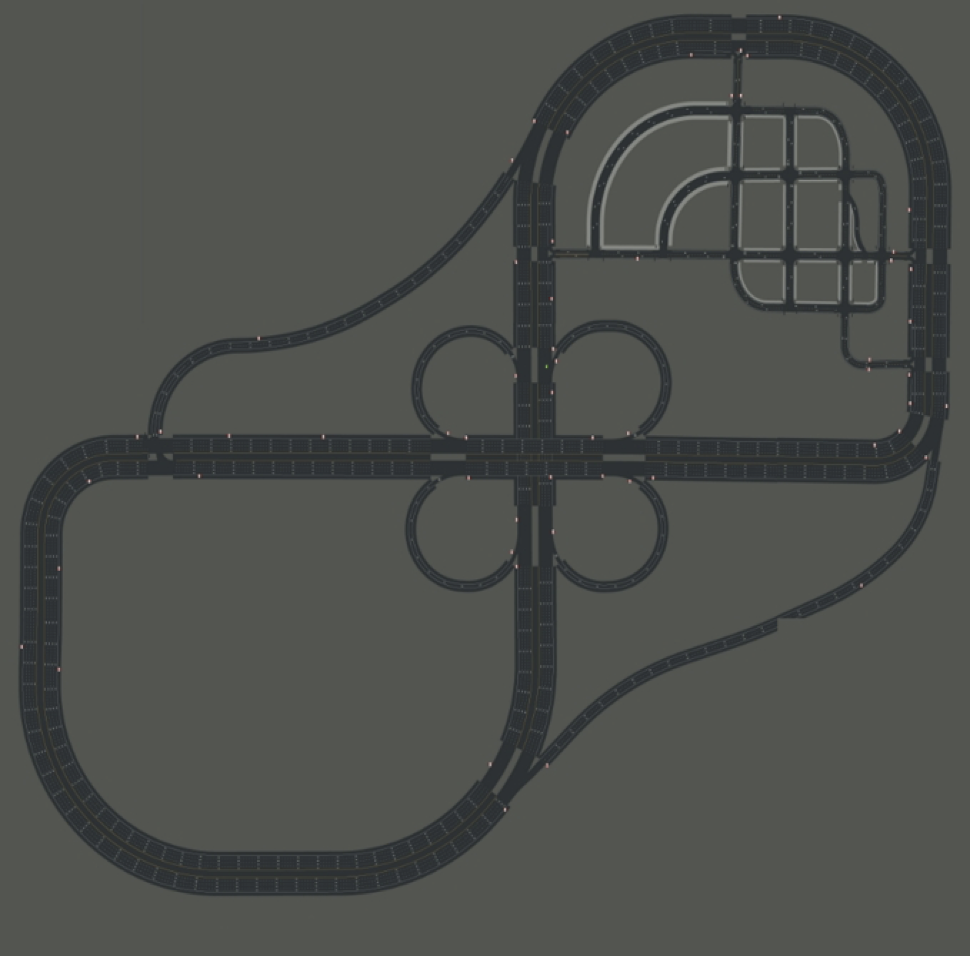}}

\caption{\footnotesize Sample egocentric view in CARLA environment on Town04 highway map. The highway layout is also shown}
\label{fig:carla}
\end{center}
\end{figure}

We replicate the autonomous driving experiments in \cite{zhang2020learning} using the CARLA simulator \cite{dosovitskiy2017carla}. The agent's goal is to drive as far as possible while avoiding collisions with other 20 vehicles in CARLA's Town04 highway map in 1000 time-steps, the reward function and environment setup are identical to \cite{zhang2020learning}. The agent has access to $5$ cameras placed on the car's roof, each covering $60^\circ$, the images are concatenated to create a single, $84\times 420$ pixel, $300^\circ$ view of the environment, sample views are shown in Figure \ref{fig:carla}. Similarly to our previous experiments, we repurpose the original training architecture, with a single shared encoder for both actor and critic, consecutive frames are stacked into a single observation, but these are processed separately, and their corresponding embeddings are concatenated; see Appendix \ref{sec:hyperparameters} for details.

Table \ref{tab:carla} summarizes the results, for these experiments DrQ was not used. Unlike the results reported in \cite{zhang2020learning}, we observe that the minor adjustments we made to the architecture are enough to fully close the performance gap between SAC and the bisimulation-based options. The lack of result differentiation in Table \ref{tab:carla} suggests that the original architecture was not well suited for the task, and bisimulation acted as a helpful regularization objective.

\begin{table}[h]
    \centering
    \scriptsize
    \begin{tabular}{ccccc}
    SAC & DBC & $\bm{\epsilon}$\textbf{-R} & PSE & $\bm{\epsilon}$\textbf{-}$\bm{\pi}$\\
    \hline
$\bm{168 \pm 	 9}$ & $160 \pm 7$ & $157 \pm 	 11$ & $158 \pm 	 11$ & $148 \pm 	 23$
\end{tabular}
\caption{Episodic reward on CARLA's Town04 highway map, $500k$ environment steps. Mean and standard deviation are computed across 5 seeds.}
\label{tab:carla}
\end{table}

\end{document}